\documentclass[10pt,twocolumn,letterpaper]{article}

\usepackage[pagenumbers]{cvpr} 

\usepackage{graphicx}
\usepackage{amsmath}
\usepackage{amssymb}
\usepackage{booktabs}
\usepackage{url}                  
\usepackage{booktabs}       
\usepackage{amsfonts}       
\usepackage{nicefrac}         
\usepackage{microtype}      
\usepackage[table,xcdraw,dvipsnames]{xcolor} 
\usepackage{graphicx,graphics}
\usepackage{caption}
\usepackage{subcaption}
\usepackage{multirow}
\usepackage{wrapfig}
\usepackage{caption}
\usepackage{listings}
\usepackage[ruled,vlined]{algorithm2e}
\usepackage{pifont}
\usepackage{amsmath}
\usepackage{array}
\usepackage{makecell}

\usepackage[pagebackref,breaklinks,colorlinks]{hyperref}
\hypersetup{
    colorlinks=true,
    linkcolor=pptred,
    filecolor=pptblue,      
    urlcolor=pptorange,
    citecolor=pptblue,
}

\usepackage[capitalize]{cleveref}
\crefname{section}{Sec.}{Secs.}
\Crefname{section}{Section}{Sections}
\Crefname{table}{Table}{Tables}
\crefname{table}{Tab.}{Tabs.}

\def\cvprPaperID{3098} 
\def\confName{CVPR}
\def\confYear{2022}

\def\method{USL\xspace}
\def\methodfive{\method\hspace{-1mm}$^\textrm{(5)}$\xspace}
\def\methodtwo{\method\hspace{-1mm}$^\textrm{(2)}$\xspace}

\newcommand{\mypar}[1]{\vspace{1mm}\noindent\textbf{#1}}
\newcolumntype{L}[1]{>{\raggedright\arraybackslash}p{#1}}
\newcolumntype{C}[1]{>{\centering\arraybackslash}p{#1}}
\newcommand{\no}{\textcolor{myred}{\ding{55}}}
\newcommand{\yes}{\textcolor{mygreen}{\checkmark}}
\def\hhline{\Xhline{2\arrayrulewidth}}

\definecolor{myred}{RGB}{220, 20, 60}
\definecolor{mygreen}{RGB}{60,179,113}

\definecolor{pptyellow}{RGB}{255, 192, 0}
\definecolor{pptblue}{RGB}{68, 114, 196}
\definecolor{pptgreen}{RGB}{112, 173, 71}
\definecolor{pptorange}{RGB}{237, 125, 49}
\definecolor{pptred}{RGB}{237, 70, 49}

\begin{document}

\title{Learning 3D Object Shape and Layout without 3D Supervision}

\author{Georgia Gkioxari{\small $^1$} \hspace{10mm} Nikhila Ravi{\small $^1$} \hspace{10mm} Justin Johnson{\small$^{1, 2}$} \\[3mm]
{\small $^1$}Meta AI \hspace{5mm} {\small $^2$}University of Michigan
} 
\maketitle

\begin{abstract}
A 3D scene consists of a set of objects, each with a shape and a layout giving their position in space.
Understanding 3D scenes from 2D images is an important goal, with applications in robotics and graphics.
While there have been recent advances in predicting 3D shape and layout from a single image, most approaches rely on 3D ground truth for training which is expensive to collect at scale.
We overcome these limitations and propose a method that learns to predict 3D shape and layout for objects without any ground truth shape or layout information: instead we rely on multi-view images with 2D supervision which can more easily be collected at scale.
Through extensive experiments on 3D Warehouse, Hypersim, and ScanNet we demonstrate that our approach scales to large datasets of realistic images, and compares favorably to methods relying on 3D ground truth.
On Hypersim and ScanNet where reliable 3D ground truth is not available, our approach outperforms supervised approaches trained on smaller and less diverse datasets. 
\footnote{Project page \url{https://gkioxari.github.io/usl/}}
\vspace{-3mm}
\end{abstract}

\section{Introduction}
\label{sec:intro}

A 3D scene consists of a set of objects,
specified by a 3D \emph{shape} for each object
and the 3D \emph{layout} of objects in space.
Understanding this 3D scene structure is critical for navigating or interacting with the world.
Unfortunately, directly measuring or perceiving 3D structure is often impractical.
For this reason, inferring the shape and layout of 3D scenes from 2D images has long been a fundamental problem in computer vision, with wide applications in robotics, autonomous vehicles, graphics, AR/VR, and beyond.

The rise of deep learning has dramatically improved 3D understanding from a single image.
Methods have advanced from estimating 3D shapes of isolated objects~\cite{choy2016r2n2,fan2017point,wang2018pixel2mesh} to predicting multiple shapes in complex scenes~\cite{gkioxari2019mesh} and even jointly predicting shape and layout~\cite{factored3dTulsiani17,Nie_2020_CVPR}.
While impressive, these methods share a flaw:
they use ground truth 3D shape and layout for training.
Creating large, varied training sets with this data is impractical,
limiting the scalability and utility of methods relying on strong 3D supervision.

Some recent approaches take an extreme position, and train on collections of images without any 3D supervision whatsoever~\cite{kanazawa2018learning,kulkarni2019csm,ucmrGoel20,kulkarni2020acsm,ye2021shelf}.
While admirable, overcoming the fundamental ambiguities of 3D from a single image requires strong category-specific shape priors, making it difficult to scale to the complexities of the real world.

\begin{figure}
    \centering
    \includegraphics[width=0.99\linewidth]{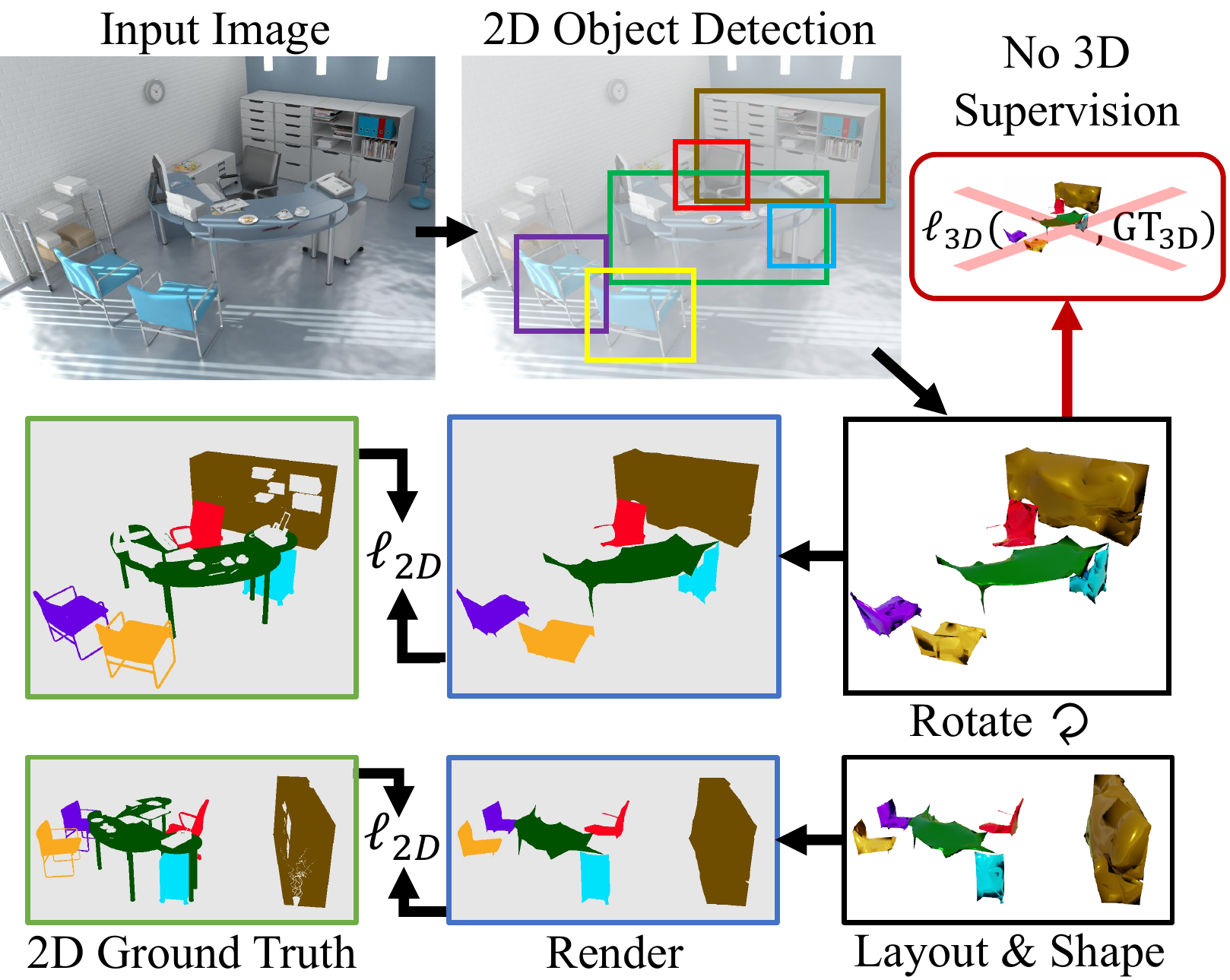}
    \vspace{-2mm}
    \caption{We propose an end-to-end model which takes an input image, detects all objects in 2D and predicts their 3D shapes and layouts. We learn from multiple 2D scene views, \eg frames of videos, and without any 3D supervision.}
    \label{fig:teaser}
     \vspace{-5mm}
\end{figure}

Another natural way to predict 3D structure is to use multiple views.
Multi-view images give weak 3D supervision and can be captured at scale using videos or multi-camera rigs.
Classical techniques such as Structure from Motion and Multi-View Stereo~\cite{hartley2003multiple} reconstruct full 3D scenes without 3D supervision, but require many views, do not predict semantics, and are not typically learned from data.

More recently, \emph{differentiable rendering} has enabled a new wave of methods that predict 3D shapes without strong 3D supervision~\cite{kato2018neural,liu2019soft,chen2019learning,ravi2020pytorch3d}.
During training a model inputs a single image and outputs a 3D shape,
which is rendered from one or more auxiliary views;
comparing the rendered prediction with 2D silhouettes in auxiliary views provides a training signal.
This pipeline is promising since it requires no ground truth 3D shapes,
instead learning solely from multi-view images and 2D image supervision which can both be collected at scale.
However, to date this technique has only been applied to simple images with a single object.

In this paper, we aim to predict 3D object shapes and layout in complex scenes from a single image, as in Fig.~\ref{fig:teaser}.
Crucially we do not use ground truth shapes or layouts during training;
instead we learn from object silhouettes in multi-view images.
We build on Mesh R-CNN~\cite{gkioxari2019mesh}, which predicts 3D shapes, but not layouts, for objects in complex scenes and relies on 3D shape supervision during training.
We augment Mesh R-CNN with a \emph{layout network} that estimates each object's 3D location, and replace expensive mesh supervision with scalable multi-view supervision.
Like prior work~\cite{kato2018neural,liu2019soft,chen2019learning,ravi2020pytorch3d} we learn via differentiable rendering and 2D losses;
however these methods only predict 3D shape -- to also predict layout we use a \emph{distance transform loss}.
We call our {\bf\underline U}nsupervised approach for {\bf\underline S}hape and {\bf\underline L}ayout estimation \method.
At test time, \method inputs a single RGB image and jointly detects objects and predicts their 3D shape and layout.

We show results on three datasets to demonstrate the utility of our scalable multi-view supervision.
First, we show results on Scene-Shapes, a synthetic dataset with scenes composed of multiple 3D Warehouse~\cite{warehouse3d} objects where our method shows strong performance compared with Mesh R-CNN trained using strong 3D shape supervision.
We then experiment on Hypersim~\cite{hypersim}, demonstrating that our approach scales to complex, realistic scenes with many objects.
Finally, we show results on ScanNet~\cite{dai2017scannet} where camera poses are estimated from BundleFusion~\cite{dai2017bundlefusion} and 2D silhouettes are estimated using PointRend~\cite{kirillov2019pointrend},
showing that we can learn from noisy real-world video without expensive ground truth.

\section{Related Work}
\label{sec:related}
\vspace{1mm}

3D scene and object reconstruction from multiple posed views has been studied extensively, from traditional Structure from Motion (SfM) and Multi-View Stereo (MVS)~\cite{hartley2003multiple, scharstein2002taxonomy}, aided by shape priors~\cite{Bao2013,blanz1999,dame2013,hane2014} to learning-based techniques~\cite{kar2017learning,kendall2017end,schmidt2017self}. 
These methods require multiple views at test time. 
In this work, we focus on single image inference.

Data driven methods predict object shape and layout from a single image of a novel scene.
\cite{gupta2013perceptual} predicts 3D object boxes from RGB-D inputs.
\cite{factored3dTulsiani17} combines oriented 3D object boxes with canonical voxel shapes from RGB inputs; Total3D~\cite{Nie_2020_CVPR} replaces voxels with meshes.
Mesh R-CNN~\cite{gkioxari2019mesh} predicts 3D object meshes via intermediate voxel predictions but does not tackle layout.
All these methods are supervised with 3D annotations, via 3D bounding boxes~\cite{Silberman:ECCV12,song2015sun} or 3D object shapes, \eg~CAD models~\cite{warehouse3d,pix3d}.
However, 3D annotations are costly and involve complicated annotation pipelines, limiting their availability to few object and scene types.
Our work shares the same goal with the above methods; we predict 3D object shapes and layouts in view coordinates from a single image, but we do so \emph{without 3D supervision}.

For the task of shape prediction, weakly supervised methods eliminate the need for 3D annotations by using category specific object priors~\cite{kanazawa2018learning,ucmrGoel20,kulkarni2019csm, kulkarni2020acsm,li2020self,zhang2020perceiving}, or 2D keypoints ~\cite{kanazawa2018learning,novotny2019c3dpo}. 
While these methods show promising results for a select few object types, their ability to scale to more classes is questionable.
In this work, we do not use object priors which allows us to scale to many more object categories.
Complementing shape, we predict object layouts, namely object positions in 3D, in an end-to-end manner.

A natural way to eliminate the need for 3D supervision and object priors is to learn from multiple views.
Differentiable rendering~\cite{loper2014opendr,kato2018neural,liu2019soft,chen2019learning,li2018differentiable,nimier2019mitsuba,ravi2020pytorch3d} allows information to flow to 3D from 2D re-projections.
\cite{kato2018neural,liu2019soft,chen2019learning,ravi2020pytorch3d,han2020drwr} achieve object reconstruction from a single view via re-projection from 2 or more views during training. 
\cite{ye2021shelf} adversarially compare silhouette re-projections with objects from an image collection.
While ground breaking, these methods focus on images of single objects in simple settings, \eg~on a white background.
In this work, we use differentiable rendering to learn from multiple views. 
However, we focus on realistic multi-object scenes, which pose significant challenges stemming from occlusion and ambiguity from multiple instances.  

Recent methods train neural networks to predict pixel-wise depth from a single image, using video frames~\cite{zhou2017unsupervised,luo2020consistent} or 3D supervision~\cite{eigen2014depth,chen2016single,li2018megadepth,yin2021learning}.
While related to layout, pixel-wise depth only captures the visible parts of the objects and commonly normalized depth is predicted. 
In this work, our goal is to reconstruct complete object shapes and predict their 3D location in metric space, \eg~in meters.

 \section{Method}
Our model inputs a single RGB image, detects objects,
and for each detected object outputs a 3D \emph{shape} (triangle mesh) and \emph{layout} (position in 3D space).
Taken together, these outputs make up a full 3D scene as shown in Figure~\ref{fig:teaser}.

During training, our model is not supervised by any ground truth 3D shape or layout information.
These annotations are expensive, so relying upon them severely limits the scale of available training data.
Instead, we use multiple RGB views of each scene together with 2D ground truth:
3D shapes predicted from one view are differentiably rendered from other views, where their 2D silhouettes are compared with 2D ground truth silhouettes.

We build on Mesh R-CNN~\cite{gkioxari2019mesh},
which extends Mask R-CNN~\cite{he2017maskrcnn} to jointly detect objects and predict 3D shapes.
We make three major modifications to Mesh R-CNN.
First, we use a new mechanism for computing vertex-aligned features called \emph{RoIMap} which better preserves aspect ratio information and improves 3D shape prediction.
Second, we introduce an additional \emph{layout head} for predicting the 3D position of each object.
Third (and most importantly), we eliminate the need for 3D shape supervision, instead learning with 2D supervision from multiple views.

\begin{figure*}
    \centering
    \includegraphics[width=0.99\textwidth]{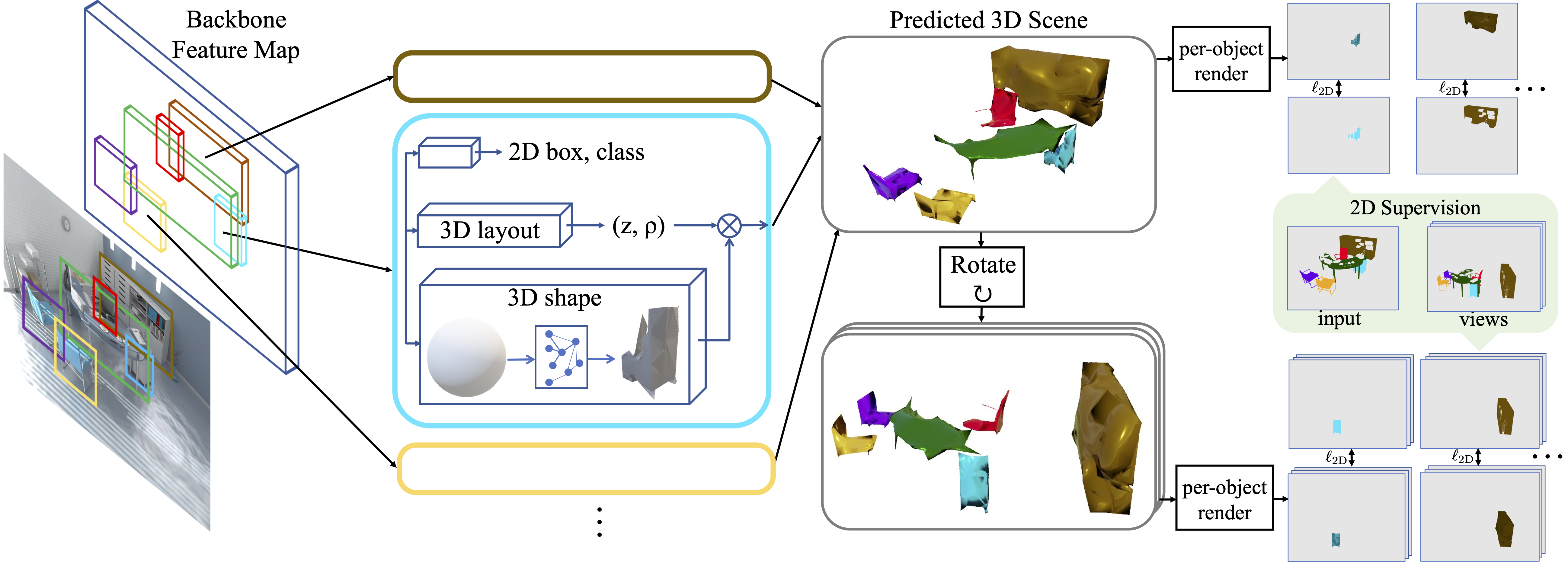}
    \vspace{-3mm}
    \caption{Our model takes as input an RGB image, detects all objects in 2D and predicts their 3D location and shape via \emph{layout} and \emph{shape} heads, respectively. The output is a scene composed of all detected 3D objects. During training, the scene is differentiably rendered from other views and compared with the 2D ground truth. We use no 3D shape or layout supervision.}
    \label{fig:overview}
    \vspace{-4mm}
\end{figure*}

\subsection{Method Overview}
The architecture of our system largely follows Mesh R-CNN~\cite{gkioxari2019mesh},
modified to allow training without ground truth shape or layout.
An overview can be seen in Figure~\ref{fig:overview}.

The input image is first processed by a \emph{backbone network} (ResNet50-FPN~\cite{he2016deep,lin2017feature} in our experiments) which extracts a \emph{backbone feature map}.
A \emph{region proposal network} (RPN)~\cite{ren2015faster} then gives category-agnostic \emph{regions of interest} (RoIs) which are processed by task-specific \emph{heads}.
The \emph{box head} performs 2D recognition; following \cite{he2017maskrcnn} it predicts a 2D bounding box and semantic category per RoI.
The \emph{layout head} performs 3D localization: for each RoI it predicts depth extent and 3D position of the object's center.
The \emph{shape head} predicts a 3D triangle mesh for each RoI; following \cite{wang2018pixel2mesh,gkioxari2019mesh} it deforms an initial sphere mesh via graph convolutions.

The box and layout heads receive input from the backbone network via \emph{RoIAlign}~\cite{he2017maskrcnn} which crops and resizes regions from the backbone feature map.
The shape head receives per-vertex features from the backbone network via \emph{RoIMap}.

During training, we assume access to $M$ views of the scene with camera poses and instance segmentations.
The model takes as input the first view and predicts the 3D scene, which is differentiably rendered from all $M$ views and compared with the 2D ground truth.

\subsection{Layout Prediction}
Our model predicts a 3D position for each object,
parameterized as an axis-aligned box
with a 3D center and length along each coordinate axis.
The box head localizes objects \emph{in the image plane};
this relies on direct image evidence, since marking the pixels belonging to each object suffices for 2D localization.
In contrast, scale/depth ambiguity makes localization \emph{vertical to the image plane} hard from image evidence alone, and must rely on prior knowledge about the world.

We thus use a separate \emph{layout head} to localize objects in depth.
It predicts each object's length $\rho$ along the depth axis and the depth $z$ of its center.
RoI features from RoIAlign~\cite{he2017maskrcnn}
 are average-pooled and passed to an MLP which predicts scalars $\tilde\rho,\tilde z\in(0, 1)$ via a sigmoid. Then
\vspace{-1mm}
\begin{equation}
  \rho = \rho_0 + \tilde\rho (\rho_1 - \rho_z)
  \hspace{2pc}
  z = z_0 + \tilde z (z_1 - z_0)
\vspace{-1mm}
\end{equation}
where $\{\rho_0,\rho_1,z_0,z_1\}$ are dataset-specific hyperparamters setting the minimum and maximum object depth and extent.
Like Mask R-CNN's box head, predictions for $\tilde\rho$ and $\tilde z$ are category-specific so the model can learn per-category priors.

\subsection{Shape Prediction}
For each detected object, the \emph{shape head} outputs a 3D triangle mesh $\mathcal{T}=(V,F)$ with vertices $V$ and faces $F$.
Predictions compose the 3D scene and are not 3D supervised.

We follow Mesh R-CNN's~\cite{gkioxari2019mesh} mesh refinement branch,
which deforms an initial mesh $\mathcal{T}_0=(V_0, F)$ via a sequence of $S$ \emph{mesh refinement stages}, each comprising three operations:
\emph{feature sampling} gives an image-aligned feature for each vertex;
\emph{graph convolution} propagates information along mesh edges;
and \emph{vertex refinement} predicts offsets $dV_i$ for each vertex and updates vertex positions $V_i=V_{i-1}+dV_i$.
The final stage's output gives the predicted shape: $V=V_S$.

Mesh R-CNN predicts a voxelized shape for each object, giving rise to instance-specific initial meshes.
This requires 3D voxel supervision and cannot be used in our setting;
we thus use an identical sphere for each object's initial mesh.

\begin{figure}
  \centering
  \includegraphics[width=0.48\textwidth]{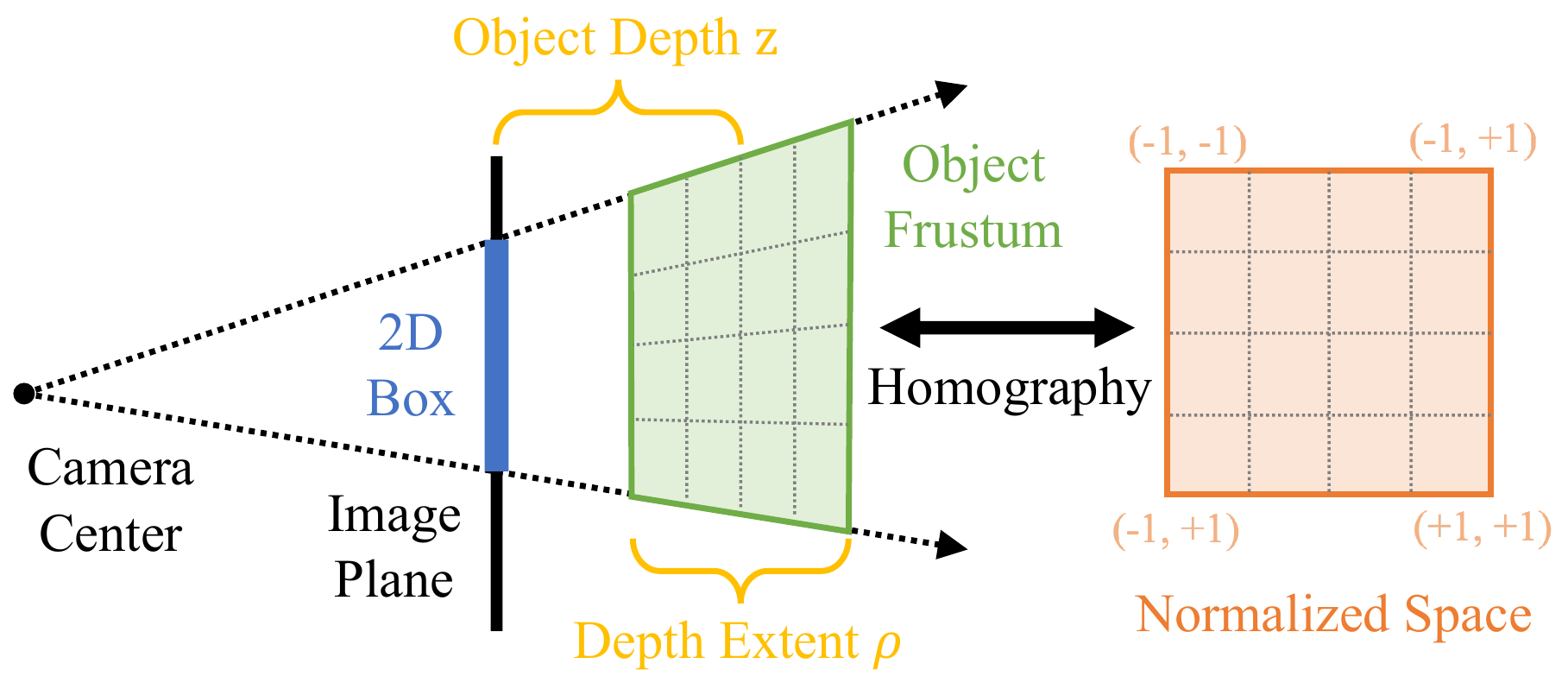}
  \vspace{-7mm}
  \caption{
    We cast shape predictions into the 3D scene by combining outputs from all heads.
    The \textcolor{pptblue}{box head} predicts a \textcolor{pptblue}{2D box} and the
    \textcolor{pptyellow}{layout head} predicts \textcolor{pptyellow}{object depth $z$}
    and \textcolor{pptyellow}{depth extent $\rho$}; together with camera intrinsics these
    define an \textcolor{pptgreen}{object frustum} in 3D space.
    The \textcolor{pptorange}{shape head} predicts a mesh in \textcolor{pptorange}{normalized space} which is mapped to the \textcolor{pptgreen}{object frustum} via a homography.
  }
  \vspace{-4mm}
  \label{fig:homography}
\end{figure}

To enable equivariance to 3D translation,
we predict shapes for each object in a normalized space with $V_0$ and each $dV_i$ in the range $[-1, +1]$.
Predicted shapes are cast into the 3D scene using a pinhole camera model:
the $[-1, +1]^3$ cube in normalized space is mapped via a homography to the object frustum defined by the camera intrinsics and the outputs of the box and layout heads (see Figure~\ref{fig:homography}).

\mypar{RoIMap.}
The shape head must precisely localize each vertex in 3D.
To this end, each mesh refinement stage receives features from the backbone by projecting the current mesh onto the image plane and bilinearly interpolating to sample a feature aligned to each vertex.
Though conceptually simple, the exact mechanism for sampling affects performance.

\begin{figure}[ht!]
\vspace{-2mm}
  \centering
  \includegraphics[width=0.85\linewidth]{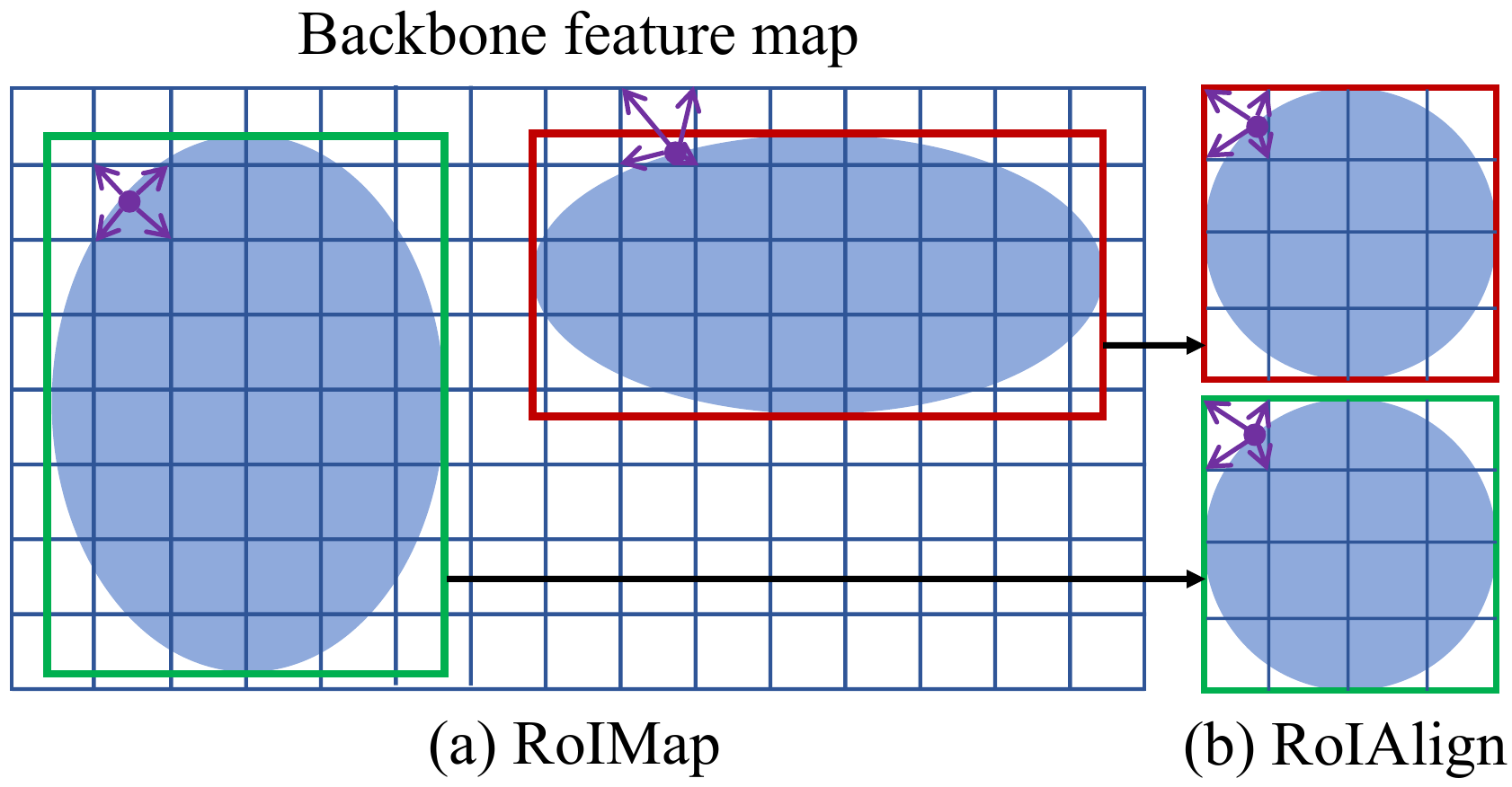}
  \vspace{-3mm}
  \caption{
    Mesh R-CNN~\cite{gkioxari2019mesh} samples vertex features from per-RoI features computed via RoIAlign.
    We instead use \emph{RoIMap} which samples vertex features directly from the backbone feature map.
  }
  \label{fig:roimap}
  \vspace{-3mm}
\end{figure}

Mesh R-CNN~\cite{gkioxari2019mesh} uses RoIAlign~\cite{he2017maskrcnn} to compute a fixed-sized feature map per RoI, then uses VertAlign to sample vertex features from the RoI features (see Fig.~\ref{fig:roimap}).
This causes several issues.
First, RoI features are a fixed square size, so vertex features do not respect the aspect ratio in the input image.
Second, repeated bilinear interpolation (first by RoIAlign then VertAlign) may cause artifacts.
Third, features cannot be computed for verticies outside the RoI.

As shown in Fig.~\ref{fig:roimap}, we overcome these issues by sampling vertex features directly from the backbone feature map rather than from RoI features.
We call this approach \emph{RoIMap}.
Our experiments in Sec.~\ref{sec:exp} show that this seemingly small change significantly improves overall performance.
A similar approach was used in \cite{kirillov2019pointrend} for instance segmentation.

\subsection{Learning without 3D Supervision}
We assume that ground truth for 3D object shapes and layouts is expensive and thus cannot be used to directly supervise the shape and layout heads.
Instead, we supervise our model using only 2D ground truth from multiple views.

During training
we sample $M$ RGB views $\{I_1, \ldots, I_M\}$ of a scene with known poses, so $R_{i\to j}$ transforms 3D points in the camera view of $I_i$ to that of $I_j$.
Let $\mathcal{O}$ be the set of objects visible in $I_1$, and $S_j^o$ be the ground-truth silhouette of $o\in\mathcal{O}$ in $I_j$.
Our model inputs $I_1$ and predicts 3D shapes $\{\mathcal{T}_1^o\}_{o\in\mathcal{O}}$ in the camera view of $I_1$.
We compute predicted 2D silhouettes from \emph{all views}, $\hat S_{1\to j}^o=\textrm{render}(R_{1\to j}\cdot \mathcal{T}_1^o)$,
using a differentiable silhouette renderer~\cite{liu2019soft,ravi2020pytorch3d}.
Our training loss for learning 3D shape and layout is then
\begin{equation}
  \small
  \mathcal{L}_\textrm{3D} = \frac{1}{|\mathcal{O}|}\frac{1}{M}\sum_{o\in\mathcal{O}}\sum_{j=1}^M\ell_\textrm{2D}(\hat S_{1\to j}^o, S_j^o)
\end{equation}
where $\ell_\textrm{2D}$ compares a pair of 2D masks using separate terms to correct errors in \emph{shape} and \emph{layout}.

\begin{figure}
  \centering
  \includegraphics[width=0.9\linewidth]{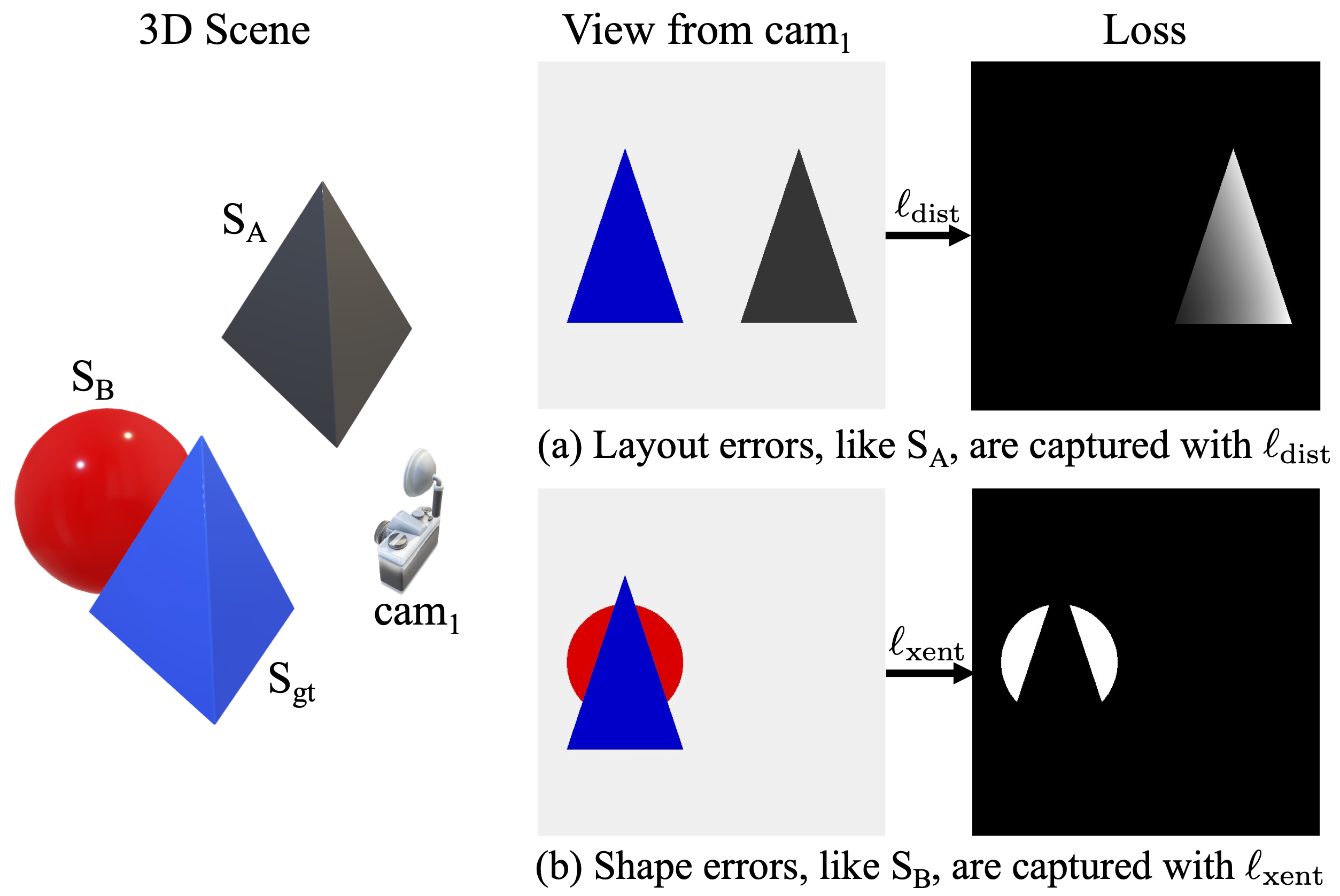}
  \vspace{-3mm}
  \caption{
    Erros in (a) layout and (b) shape dictate the appropriate loss functions
    for comparing image silhouettes.
  }
  \label{fig:losses}
  \vspace{-4mm}
\end{figure}

As shown in Fig.~\ref{fig:losses}, a pixel-wise cross-entropy loss $\ell_\textrm{xent}$ gives a useful learning signal when two masks overlap but differ in \emph{shape}.
However when there is no overlap $\ell_\textrm{xent}$ equally penalizes all predictions and thus does not tell the model how to correct errors in \emph{layout}.

We thus introduce a \emph{distance transform loss} $\ell_\textrm{dist}$ which penalizes 2D distances between masks:
\begin{equation}
  \label{eq:l_dist}
  \small
  \ell_\textrm{dist}(\hat S, S)
    = \int_{\hat S}\inf_{s\in S}\|\hat s-s\|_2^2\;d\hat s
    + \int_{S}\inf_{\hat s\in\hat S}\|\hat s-s\|_2^2\;ds.
\end{equation}
This loss penalizes non-overlapping silhouettes
depending on their screen-space distance, aiding layout prediction.
We approximate $\ell_\textrm{dist}$ as a bidirectional Chamfer distance between points sampled from $\hat S$ and $S$.
To sample from $\hat S$, we sample from the surface of the mesh (like \cite{smith2019geometrics}) then project onto the image plane,
so $\ell_\textrm{dist}$ does not require computing $\hat S$.

Distance transforms have a long history in object detection~\cite{felzenszwalb2009object,felzenszwalb2012distance};
a similar loss was used by \cite{kanazawa2018learning} to learn texture.

Our overall loss on pairs of 2D silhouettes is thus
\begin{equation}
  \small
  \ell_\textrm{2D}(\hat S, S) = \ell_\textrm{dist}(\hat S, S) + \mathbf{1}[IoU(\hat S, S){>}0.5] \cdot \ell_\textrm{xent}(\hat S, S)
\end{equation}
only applying $\ell_\textrm{xent}$ for overlapping silhouettes (IoU${>}0.5$).

Our full training loss is a weighted combination of Mask R-CNN's 2D losses,
our 3D loss $\mathcal{L}_\textrm{3D}$, and 3D shape regularizers encouraging smooth mesh predictions.

\mypar{Dynamic rendering.}
Objects in real-world scenes tend to occupy few image pixels,
so naively computing $\hat S^o_{1\to j}$ spends significant resources rasterizing pixels not occupied by objects, limiting rendering resolution.
We thus use a \emph{dynamic rendering} scheme: when computing $\hat S^o_{1\to j}$, we only render a region which is the union of the ground truth silhouette and the projection of the predicted mesh $\mathcal{T}^o_1$ onto view $j$.
This allows rendering at 4$\times$ resolution vs naive rendering, which captures finer object details and improves results.

\begin{table}
  \centering
  \scalebox{0.99}{
  \begin{tabular}{L{.34\linewidth}|C{.13\linewidth}C{.08\linewidth}C{.1\linewidth}C{.1\linewidth}}
  	\toprule
	\multicolumn{1}{c}{}    & \multicolumn{2}{c}{3D Metrics} & \multicolumn{2}{c}{Mask$_\textrm{2D}$ IoU}  \\
	Model  & Ch.($\downarrow$)& F$_1$  & Input & Views \\
	\hhline
	Fixed depth & 0.275 & 20.1 & 18.4 & 14.1   \\ 
	Random depth & 0.202 & 23.3 & 21.5 & 17.6 \\ 
	\methodtwo & 0.050 & 62.9 & \bf{53.6} & 43.4 \\ 
	\methodfive & \bf{0.034} & \bf{70.9} & 52.3 & \bf{46.7} \\ 
	\hline
	\methodfive w/o RoIMap & 0.059 & 55.1 & 51.1 & 40.9 \\ 
	\methodfive w/o $\ell_\textrm{dist}$ & 0.039 & 68.9 & 42.9 & 37.1 \\ 
	\hline
	Mesh R-CNN~\cite{gkioxari2019mesh} & \textcolor{myred}{\bf{0.015}} &  \textcolor{myred}{\bf{87.9}} & \textcolor{myred}{\bf{61.5}} &  \textcolor{myred}{\bf{57.7}} \\ 
  \end{tabular}
  } 
  \vspace{-3mm}
   \caption{Performance on Scene-Shapes val. We report a \emph{random} and a \emph{fixed} depth baseline which place a sphere for each object at random and a fixed depth, respectively, We report our model, \method, trained with 2 \& 5 views and ablate RoIMap and $\ell_\textrm{dist}$. We compare to Mesh R-CNN~\cite{gkioxari2019mesh} which is the supervised state-of-the-art.}
   \label{tab:multishapenet_abl}
   \vspace{-3mm}
\end{table}

\section{Experiments}
\label{sec:exp}

We experiment on three datasets: Scene-Shapes, Hypersim~\cite{hypersim} and ScanNet~\cite{dai2017scannet}.
Scene-Shapes forms simple scenes from 3D Warehouse~\cite{warehouse3d} objects, while Hypersim and ScanNet contain video sequences of complex scenes with multiple objects under varying appearance, occlusion and lighting conditions; a stark difference to single object benchmarks.

Scene-Shapes provides ground truth 3D shape and layout, enabling comparison with supervised methods and the use of 3D evaluation metrics.
3D ground truth is not available on Hypersim and ScanNet, so we resort to proxy metrics by comparing rendered predictions to 2D ground truth from multiple views.
We perform extensive quantitative analysis and show predictions on challenging images of novel scenes.
Compared to state-of-the-art supervised methods trained on smaller and less diverse 3D annotated datasets,
we show that our method can better generalize to novel scenes.

\mypar{Metrics.}
In the absence of 3D ground truth, we use an evaluation scheme which relies on multi-view 2D comparisons.

Specifically, we project each object's predicted 3D shape to all available views of the scene.
In each view we compute the intersection-over-union (IoU) between the rendered prediction and ground truth object mask in that view.
We report two metrics: \emph{Mask$_\textrm{2D}$ IoU input} is the IoU in the view the model receives as input,
and \emph{Mask$_\textrm{2D}$ IoU views} is the mean across all other views;
both are averaged over all scenes.

\begin{figure}[t]
  \centering
  \scalebox{0.99}{
    \begin{tabular}{C{.3\linewidth}C{.31\linewidth}C{.31\linewidth}}
    \hspace{-8mm} Input Image & \hspace{-12mm}\method & \hspace{-12mm}Mesh R-CNN~\cite{gkioxari2019mesh}
    \end{tabular}
    } 
   \includegraphics[width=0.99\linewidth]{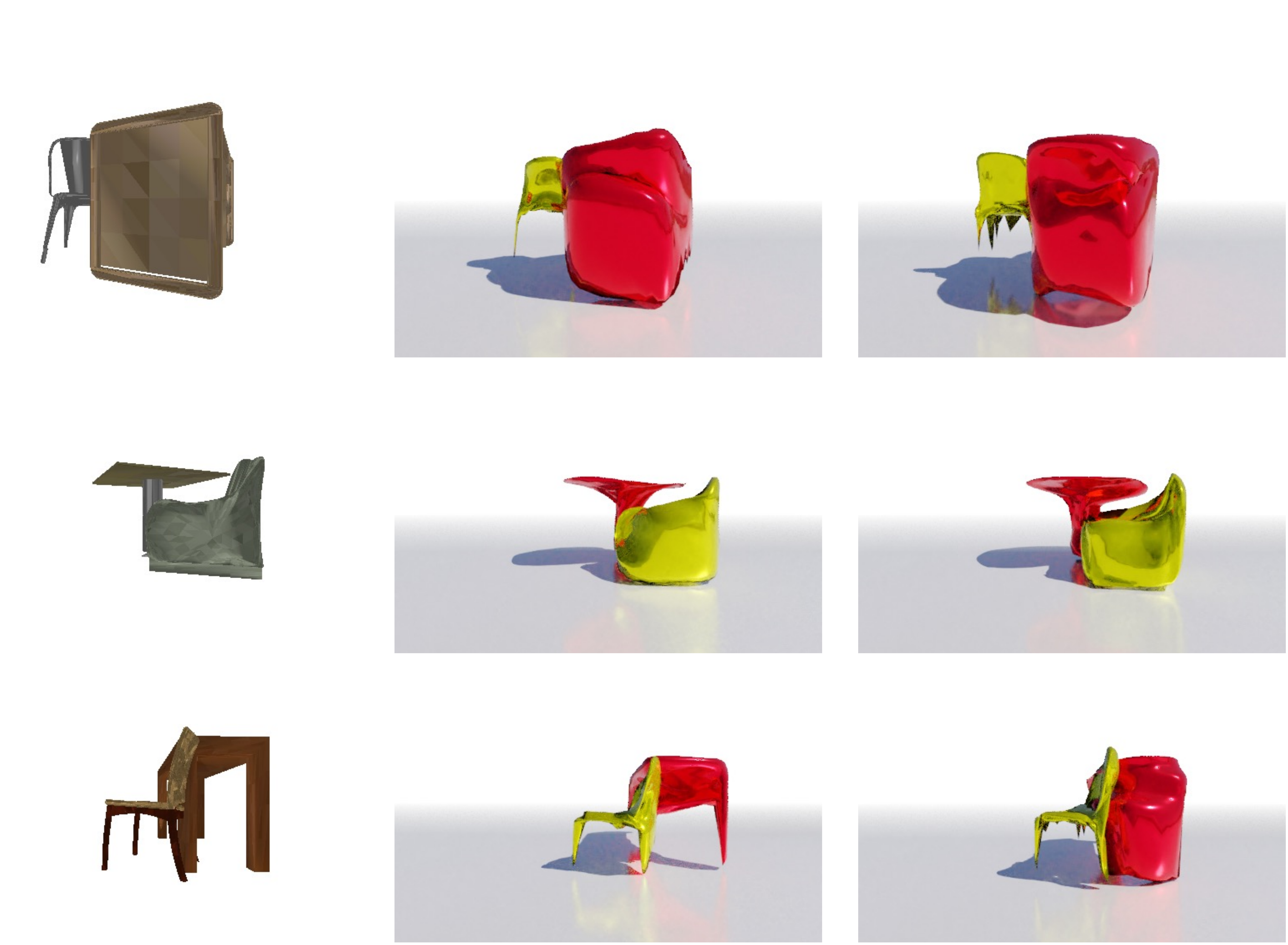}
   \caption{Predictions on Scene-Shapes val. We show the input image (left) and the predicted 3D objects and layout of our \method (middle) and 3D supervised Mesh R-CNN~\cite{gkioxari2019mesh} (right).}
   \label{fig:multishapenet_preds}
   \vspace{-1mm}
\end{figure}

\subsection{Results on Scene-Shapes}
We introduce Scene-Shapes, a dataset of scenes formed of 3D Warehouse~\cite{warehouse3d} objects.
Its scenes contain object pairs from three object types, namely \emph{chair, sofa} and \emph{table}.
Objects are placed at random 3D locations and poses and scenes are rendered from multiple viewpoints.
The dataset consists of 86.4k images and 4k unique object shapes, split into 80\%/10\%/10\% for train/test/val.
\emph{Each split contains unique object models and scenes}.
See Appendix~\ref{app:multishapenet} for more details.
This dataset provides 3D ground truth, so we can evaluate predicted shapes and layouts in 3D.

\mypar{Training details.} 
We follow Mesh R-CNN~\cite{gkioxari2019mesh} and train using Adam for 25 epochs with batches of 64 images on 8 V100 GPUs.
For each example in the batch, we randomly sample $M$ views from the corresponding scene.
Input images are 512$\times$512; we render with PyTorch3D~\cite{ravi2020pytorch3d} at a resolution of 128$\times$128 with 10 faces per pixel; blur radius and blend sigma are $10^{-3}$.
The backbone ResNet50 is pretrained on ImageNet; other parameters are learned from scratch.

\mypar{Evaluation.}
We evaluate on val, which contains objects and scenes disjoint from train.
Since 3D ground truth is available, in addition to Mask$_\textrm{2D}$ IoU we also report standard 3D metrics:
3D chamfer distance and F$_1@0.1$m, following~\cite{gkioxari2019mesh}.

\begin{figure*}[t]
  \centering
   \includegraphics[width=0.95\linewidth]{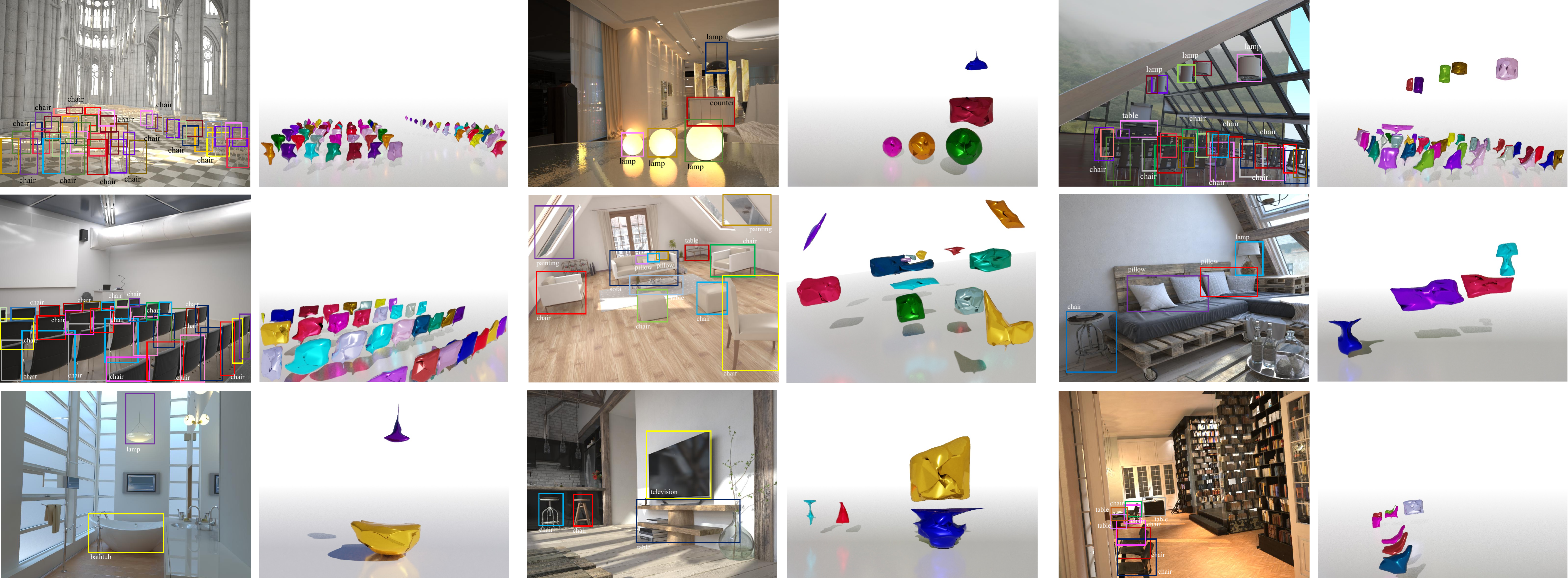}
   \vspace{-2mm}
   \caption{Predictions on Hypersim. For each example, we show the input image and the detected 2D objects (left) and the predicted 3D objects and layout (right). Examples are of complex scenes with many objects of diverse appearances and types. See our video animations.}
   \label{fig:hypersim_preds}
   \vspace{-4.2mm}
\end{figure*}

\mypar{Results}
are shown in Table~\ref{tab:multishapenet_abl}.
We compare to variants of \method with $M{\in}\{2,5\}$ views and ablate RoIMap and the distance transform loss, $\ell_{\hspace{0.3mm} \textrm{dist}}$. 
We report a \emph{random} baseline which predicts each object as a sphere with random depth $z\in[1.4, 2.0]$ and depth extent $\rho\in[0.1,1.0]$,
and a \emph{fixed-depth} baseline which predicts each object as sphere with $z=1.7,\rho=1.0$.
We also compare to Mesh R-CNN~\cite{gkioxari2019mesh} trained on Scene-Shapes with full 3D supervision.

From Table~\ref{tab:multishapenet_abl}, Mesh R-CNN performs the best as is expected (7$^\textrm{th}$ row).
Our \methodfive performs best among all unsupervised baselines (4$^\textrm{th}$ row).
We observe that performance drops when we replace RoIMap with RoIAlign (5$^\textrm{th}$ row) and when we omit $\ell_{\hspace{0.3mm} \textrm{dist}}$ (6$^\textrm{th}$ row).
A model trained with 2 views (3$^\textrm{rd}$ row) performs worse than the 5-view model.
Finally, we note that our Mask$_\textrm{2D}$ IoU on views correlates with 3D metrics, validating its choice as a proxy for 3D performance.  

Figure~\ref{fig:multishapenet_preds} compares predictions from \methodfive and Mesh R-CNN on Scene-Shapes.
Though \method receives no 3D supervision, it predicts accurate 3D layouts and shapes even when objects are occluded. 
Mesh R-CNN makes more accurate predictions (see Table~\ref{tab:multishapenet_abl}) but requires 3D supervision, which is expensive to obtain at scale.
See Appendix~\ref{app:multishapenet} for more qualitative examples.

\subsection{Results on Hypersim}
We experiment on Hypersim~\cite{hypersim}, a dataset of 461 complex scenes,
each rendered along camera trajectories giving 77,400 images
with ground truth pose, semantic and instance masks for 40 object categories,
and instance IDs linking objects across views.
Hypersim has an average of 50 instances and 10 object types per image;
in contrast, COCO~\cite{lin2014microsoft} images have just 7 instances on average.

\mypar{Training details.}
We train on Hypersim's train split of 365 scenes.
We follow the Mask R-CNN~\cite{he2017maskrcnn} recipe~\cite{wu2019detectron2},
training with batches of 16 images on 8 V100 GPUs for 80k iterations.
We use SGD with initial learning rate $10^{-2}$, decaying by $0.1$ after 66k and 74k iterations.
For each example in the batch, we randomly sample $M$ views from the corresponding video.
We use PyTorch3D~\cite{ravi2020pytorch3d} and render dynamically at 72$\times$72, 10 faces per pixel, blur radius and blend sigma of $10^{-3}$.
The backbone ResNet50-FPN is pretrained on COCO; all other parameters are learned from scratch.
More details can be found in Appendix~\ref{app:hypersim}.

\begin{table*}[h!!]
  \centering
  \scalebox{0.99}{
  \begin{tabular}{L{.28\linewidth}|C{.07\linewidth}C{.05\linewidth}|C{.06\linewidth}C{.06\linewidth}C{.06\linewidth}C{.06\linewidth}C{.06\linewidth}C{.06\linewidth}}
  	\toprule
	\multicolumn{1}{c}{}   & \multicolumn{2}{c}{Predicts}   & \multicolumn{2}{c}{Box$_\textrm{2D}$ gIoU} & \multicolumn{2}{c}{Mask$_\textrm{2D}$ IoU} & \multicolumn{2}{c}{Depth $L_1$ ($\downarrow$)}  \\
	Model   & Layout & Shape  & Input & Views  & Input & Views & Input & Views \\
	\hhline
	Random depth & \no & \no & 1.00 & 0.13 & 0.58 & 0.20 & 3.55 & 3.28\\ 
	Fixed depth & \no & \no & 1.00 & 0.22 & 0.58 & 0.24 & 2.73 &  2.59 \\ 
	Layout-Only\hspace{-0.1mm}$^\textrm{(2)}$ & \yes & \no & 1.00 & 0.25 & 0.58 & 0.25 & 2.51 & 2.35 \\ 
	Layout-Only\hspace{-0.1mm}$^\textrm{(5)}$ & \yes & \no & 1.00 & \bf{0.37} & 0.58 & 0.30 &1.81 & 1.74  \\ 
	\methodtwo &\yes & \yes & 1.00 & 0.30 & 0.73 & 0.33 & 1.90 & 1.83 \\ 
	\methodfive & \yes & \yes & 1.00 & 0.33 & \bf{0.74} & \bf{0.34} & \bf{1.78} & \bf{1.72} \\ 
	\hline
	Mesh R-CNN~\cite{gkioxari2019mesh} + \methodfive layout & \yes & \yes & 1.00 & 0.21 & 0.36 & 0.22 & \bf{1.78} & \bf{1.72} \\ 
	\hline
	\methodfive w/ detections & \yes & \yes & 0.92  & 0.31 & 0.72 & 0.33 & 1.80 & 1.74  \\ 
  \end{tabular}
  } 
  \vspace{-2mm}
   \caption{Results on Hypersim val. We report a \emph{random} and a \emph{fixed-depth} baseline which place a sphere at random and a fixed depth, respectively, for each object. We train \emph{Layout-Only} variants of our approach with 2 and 5 views which represent objects as spheres at the predicted 3D locations. Finally, we train our \method with 2 and 5 views which learn both shape and layout. We report the performance of Mesh R-CNN~\cite{gkioxari2019mesh} pretrained on Pix3D~\cite{pix3d} for 3D shape prediction and combined with \methodfive layout predictions. The final row shows the performance of \methodfive when using the model's object predictions instead of ground truth detections.}
   \label{tab:hypersim_abl}
   \vspace{-3mm}
\end{table*}

\begin{table*}[h!!]
  \begin{subtable}{0.4\linewidth}
  \small{
  \begin{tabular}{L{.18\linewidth}|C{.1\linewidth}C{.1\linewidth}C{.1\linewidth}C{.1\linewidth}}
  	\toprule
	\multicolumn{1}{c}{}  & \multicolumn{2}{c}{Mask$_\textrm{2D}$ IoU} & \multicolumn{2}{c}{Depth $L_1$}  \\
	 & Input & Views  & Input & Views \\
	\hhline
	RoIAlign & 0.67 & 0.20 & 4.38 & 3.86                          \\ 
	RoIMap & \bf{0.74} & \bf{0.34} & \bf{1.78} & \bf{1.72}  \\ 
  \end{tabular}
  \caption{}
  \label{tab:hypersim_roi}
  } 
  \end{subtable}
  \hspace{-0.5cm}
  \begin{subtable}{0.35\linewidth}
  \small{
  \begin{tabular}{C{.05\linewidth}|C{.13\linewidth}C{.13\linewidth}C{.13\linewidth}C{.13\linewidth}}
  	\toprule
	\multicolumn{1}{c}{}  & \multicolumn{2}{c}{Mask$_\textrm{2D}$ IoU} & \multicolumn{2}{c}{Depth $L_1$} \\
	$\ell_\textrm{dist}$ & Input & Views  & Input & Views \\
	\hhline
	\no              & 0.74 & 0.28 & 2.61 & 2.44                      \\ 
	\yes & 0.74 & \bf{0.34} & \bf{1.78} & \bf{1.72}   \\ 
  \end{tabular}
  \caption{}
  \label{tab:hypersim_dist}
  } 
  \end{subtable}
  \hspace{-0.2cm}
  \begin{subtable}{0.24\linewidth}
  \small{
  \begin{tabular}{L{.4\linewidth}|C{.2\linewidth}C{.2\linewidth}}
  	\toprule
	\multicolumn{1}{l}{}  & \multicolumn{2}{c}{Mask$_\textrm{2D}$ IoU} \\
	Model & Input & Views \\
	\hhline
	Sphere-Only   & 0.58 & 0.45               \\ 
	\methodfive     & \bf{0.74} & \bf{0.53}   \\ 
  \end{tabular}
  \caption{}
  \label{tab:hypersim_gtZ}
  } 
  \end{subtable}
  \vspace{-3mm}
  \caption{Ablations on Hypersim for (a) \emph{RoIAlign} vs. \emph{RoIMap}, (b) distance transform loss $\ell_\textrm{dist}$, and with (c) oracle depth.}
  \vspace{-4mm}
\end{table*}


\mypar{Evaluation.}
We evaluate on Hypersim val, which consists of 46 scenes disjoint from train.
Hypersim does not provide ground truth 3D object shape and layout information so we cannot report 3D metrics. 
However, it provides ground truth pixel-wise metric depth (in meters) which when combined with instance masks gives ground truth depth on the visible parts of each object.
This allows us to additionally report a \emph{Depth $L_1$} metric (for both the input and other views) between the true and predicted nearest depth of each object.
Finally, we also report \emph{Box$_\textrm{2D}$ gIoU} on input and views, by computing the 2D image-aligned boxes bounding the predicted and true object silhouettes, used in \emph{Mask$_\textrm{2D}$ IoU}, and measure the gIoU~\cite{Rezatofighi_2018_CVPR}, a generalization of traditional IoU which measures the proximity between two boxes.

\mypar{Results.}
We compare to a \emph{random} baseline which predicts each object as a sphere at random depth $z{\in}[1.0, 10.0]$ and depth extent $\rho{\in}[0.1, 1.0]$,
and a \emph{fixed-depth} baseline, which predicts objects as spheres fixed at $z{=}5$, $\rho{=}0.5$.
We train \emph{layout-only} variants with $M{\in}\{2,5\}$ views, which learn \emph{layout} but predict \emph{shape} as fixed spheres.
We train \method with $M{\in}\{2,5\}$ views for both 3D layout and shape.

Finally, we compare to Mesh R-CNN~\cite{gkioxari2019mesh}.
Hypersim does not release public 3D shape annotations needed to train Mesh R-CNN,
so we instead train Mesh R-CNN on Pix3D as in \cite{gkioxari2019mesh}.
Pix3D only provides ground truth \emph{shape} but not \emph{layout}, so Mesh R-CNN trained on Pix3D can only predict shape.
We thus combine \emph{shapes} predicted by Mesh R-CNN with \emph{layouts} predicted by \method for this baseline.

Table~\ref{tab:hypersim_abl} shows the performance on val.
To ensure fair comparisons we decouple 3D understanding from 2D detection by using ground truth 2D boxes on the input image during evaluation for all baselines (Box$_\textrm{2D}$ gIoU = 1.0 for input). 
Notably, our model's 2D object detector, trained jointly with the shape and layout networks, achieves an AP of 64\% and AP$^\textrm{50}$ of 73\%.
We report our model's performance when using its own object detections in the last row of Table~\ref{tab:hypersim_abl}.

From Table~\ref{tab:hypersim_abl},
our \methodfive model (6$^\textrm{th}$ row) outperforms the \emph{layout-only}, \emph{random} and \emph{fixed-depth} baselines for Mask$_\textrm{2D}$ IoU and Depth $L_1$.
The \emph{layout-only}$^\textrm{(5)}$ baseline has a higher Box$_\textrm{2D}$ gIoU but a lower Mask$_\textrm{2D}$ IoU on views than \methodfive (4$^\textrm{th}$ vs. 6$^\textrm{th}$ row), indicating that it works well for layout but not for shape. 
Training with $5$ views is better than $2$ views (5$^\textrm{th}$ vs. 6$^\textrm{th}$ and 3$^\textrm{rd}$ vs. 4$^\textrm{th}$ row), which is expected as 5 views during training provide more information. 
Using more than 5 views does not improve performance further, likely because walk-through videos cap the number of frames with new information about each part of the scene.
The Mesh R-CNN baseline achieves low performance despite being supervised, proving that existing 3D annotated datasets are insufficient and don't generalize well to more complex scenes.
Finally,  we observe that Mask$_\textrm{2D}$ IoU correlates with Depth $L_1$ on views, as models with higher IoU have lower depth error.
In the absence of any 3D ground truth, Mask$_\textrm{2D}$ IoU is likely to serve as a good proxy for 3D metrics.
 
Table~\ref{tab:hypersim_roi} compares RoIMap to RoIAlign and shows the impact of RoIMap on the performance.
Table~\ref{tab:hypersim_dist} ablates the distance transform $\ell_\textrm{dist}$ (Equation~\ref{eq:l_dist}) which proves crucial to our model's performance.
Finally, note that Mask$_\textrm{2D}$ IoU on views captures both shape and layout errors; wrong layout predictions even for accurate shapes can result in low IoU. 
To decouple performance for layout and shape, we compare our \methodfive to a \emph{sphere-only} baseline, which represents each object as a sphere, and provide true object depth for both models at test time in Table~\ref{tab:hypersim_gtZ}.
From Table~\ref{tab:hypersim_gtZ} we see that we outperform \emph{sphere-only} for shape. 

Figure~\ref{fig:hypersim_preds} shows predictions on Hypersim for diverse novel scenes with many object instances and types, including \emph{lamp}, \emph{painting}, \emph{sofa}, \emph{chair}, \emph{table}, \emph{tv}, \emph{bathtub} and \emph{counter}.
We observe that our model captures layout well, while object shapes are roughly correct but certainly less refined.
Predicting detailed 3D shapes without 3D supervision is hard.
In addition to the lack of 3D supervision, we learn from views extracted from walk-through videos which capture scenes from a constrained, far from 360$^o$, set of views (\eg backs of couches are never seen, etc.).
This is in contrast to Scene-Shapes, where 360$^o$ scene views are available, and thus our model is able to capture shape more accurately.

\mypar{Comparison to Total3D.}
We compare to Total3D~\cite{Nie_2020_CVPR}, a state-of-the-art fully supervised method for predicting shape and layout from a single image.
Total3D learns a \emph{layout model} on SUN~RGB-D~\cite{song2015sun} which provides oriented 3D object bounding boxes, and learns a \emph{shape model} on Pix3D~\cite{pix3d} which provides image aligned CAD models for 9 object classes.
At test time, predictions from the shape model are positioned according to predictions from the layout model; this gives final predictions in \emph{view coordinates}.

Figure~\ref{fig:comparison} qualitatively compares to Total3D on randomly selected input images; see more in Appendix~\ref{app:total3d}.
Despite being supervised, we observe that Total3D tends to fetch the nearest shape for the object class, which does not match the object's appearance in the input. 
For example, in the 1$^\textrm{st}$ example it predicts a rectangular table instead of a round one shown in the image.
Regarding layout, Total3D struggles to place objects in the correct relative locations, resulting in large shape intersections and erroneous layouts.
We also note that the 3D objects do not align with the 2D objects (2$^\textrm{nd}$ vs 4$^\textrm{rd}$ col); Total3D does not enforce alignment with 2D contrary to our approach which does so by design (Figure~\ref{fig:homography}).
Figure~\ref{fig:comparison} proves that Total3D has an extremely hard time generalizing to complex scenes, despite being supervised with 3D ground truth for shape and layout.
We draw the same conclusion when comparing to Mesh R-CNN in Table~\ref{tab:hypersim_abl}. 
This is further proof that existing 3D annotated datasets, like Pix3D and SUN RGB-D, are not adequate.
Training on larger more diverse datasets could improve Total3D's performance, but 3D annotations are expensive to collect at large scales. 
Our approach is a first attempt to tackle 3D object layout on complex scenes bypassing the need for 3D supervision and can be scaled naturally to more train data.

\begin{figure}[t!]
  \centering
  \scalebox{0.8}{
    { \small
    \begin{tabular}{C{.2\linewidth}C{.4\linewidth}C{.4\linewidth}}
    \hspace{-9mm} Input Image & \hspace{-1.5mm} Total3D~\cite{Nie_2020_CVPR} & \hspace{3mm} \method \\
    \end{tabular}
    } 
    } 
  \includegraphics[width=1.0\linewidth]{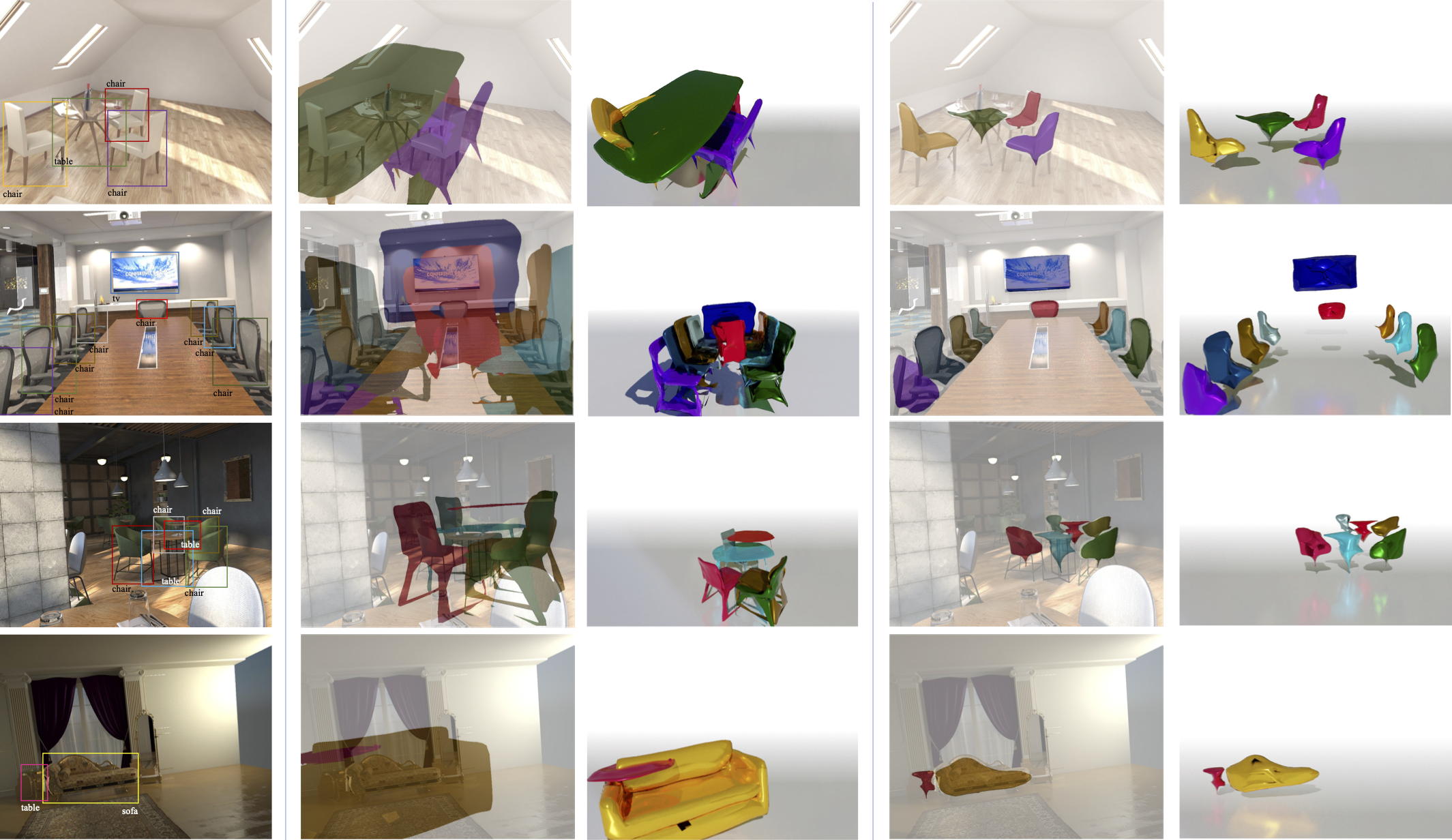}
  \vspace{-6mm}
   \caption{Comparison of Total3D~\cite{Nie_2020_CVPR} and our approach. The input image is shown in the 1$^\textrm{st}$~col. We show the predicted 3D shapes and layout perspectively projected into the image plane along with a 3D visualization of the predicted layout, for Total3D (2$^\textrm{nd}$ \& 3$^\textrm{rd}$~col) and our approach (4$^\textrm{th}$ \& 5$^\textrm{th}$~col). More in Appendix~\ref{app:total3d}.}
   \label{fig:comparison}
   \vspace{-3mm}
\end{figure}

\subsection{Results on ScanNet}
We experiment on ScanNet~\cite{dai2017scannet}, a dataset of videos of indoor scenes with reconstructed cameras.
We detect object instances and object tracks by applying a PointRend~\cite{kirillov2019pointrend} model pre-trained on COCO~\cite{lin2014microsoft}.
COCO consists of 80 object categories out of which more than 30 are detected on ScanNet, including \emph{bottle}, \emph{keyboard}, \emph{books}, \emph{potted plant}, \emph{tv} and more.
We track objects with a simple mask IoU heuristic and keep tracks of at least 50 frames to ensure diversity of object views for training and evaluation.
This results in 210k instances across 160k frames and more than 2700 tracks.

\mypar{Training details.} 
We train on the ScanNet training set for 200k iterations, with a learning rate of $0.01$ which drops by $0.1$ at 150k \& 170k iterations. 
During training, we sample $M$ views from the object tracks with a probability proportional to the time distance from the input to encourage diverse views.
We initialize the backbone with COCO-pretrained weights, and learn the remaining parameters from scratch.

\mypar{Results.}
Table~\ref{tab:scannet_abl} shows results on ScanNet val, which consists of scenes distinct from train. 
We evaluate a \emph{random} and \emph{fixed-depth} baseline and our \method trained with $5$ views.\linebreak
We report Box$_\textrm{2D}$ gIoU and Mask$_\textrm{2D}$ IoU for the input and views.
For fair comparisons, we feed the same input boxes detected from PointRend to all models.
For the view metrics, we evaluate on the furthest $20\%$ frames in the track from each input view to put emphasis on diverse object views.
From Table~\ref{tab:scannet_abl}, our model outperforms all baselines.

\begin{figure}
  \centering
  \includegraphics[width=0.95\linewidth]{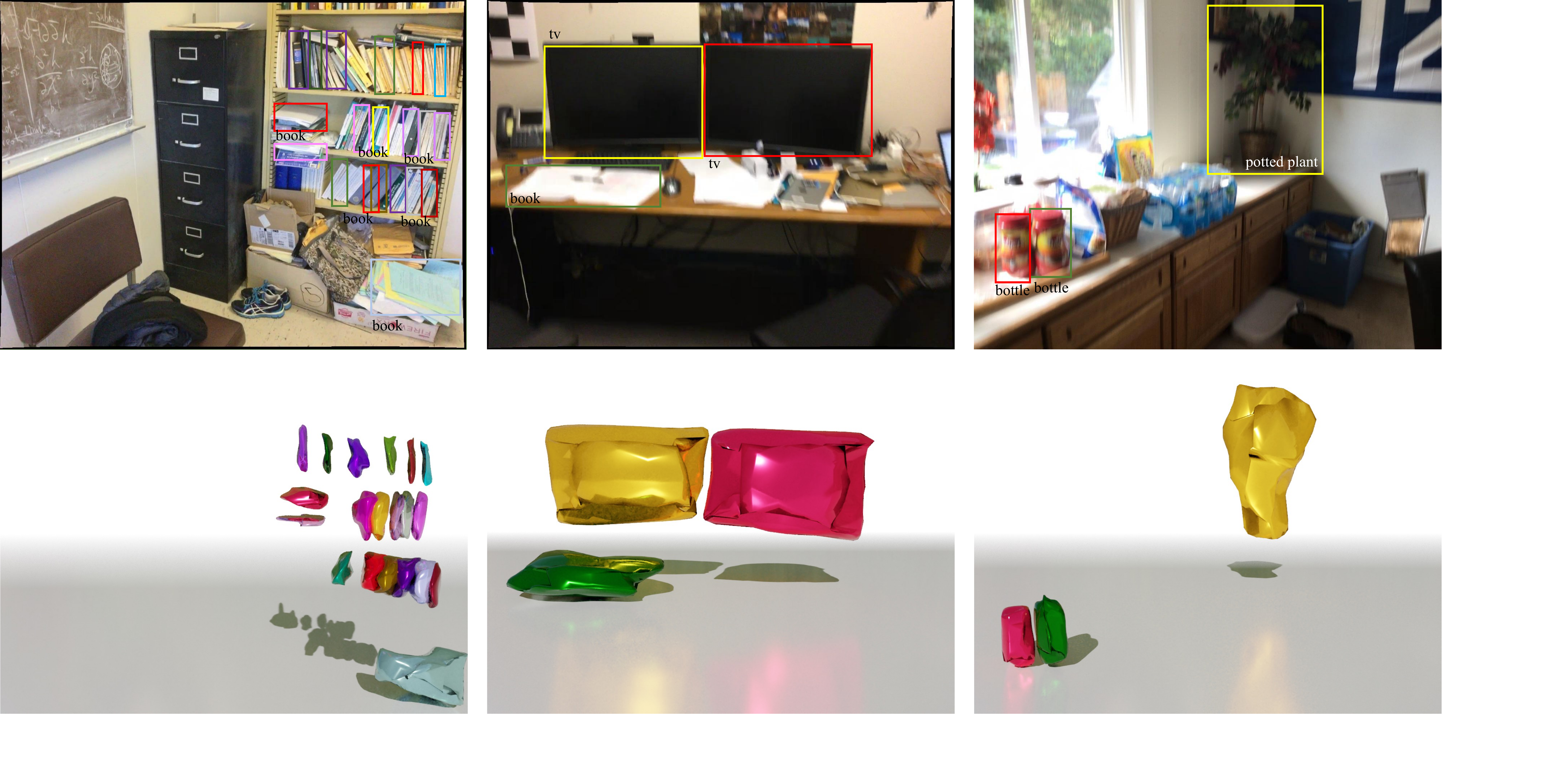}
  \vspace{-3mm}
   \caption{Predictions on ScanNet val. We show the input image along with detected 2D objects (top) and the predicted 3D objects and layout from our approach (bottom).}
   \label{fig:scannet_preds}
   \vspace{-3mm}
\end{figure}

Figure~\ref{fig:scannet_preds} shows predictions on ScanNet.
We show the input image with 2D object detections superimposed (top) and the predicted 3D object layouts (bottom) for a variety of scenes.
ScanNet is challenging as frames have motion blur and are narrow views of scenes with heavy object 
truncations by the image border. 
Yet, our model is able to reason about 3D objects and their location in the scene.

\begin{table}
  \centering
  \scalebox{0.95}{
  \begin{tabular}{L{.25\linewidth}|C{.04\linewidth}C{.04\linewidth}|C{.08\linewidth}C{.08\linewidth}C{.08\linewidth}C{.08\linewidth}}
  	\toprule
	\multicolumn{1}{c}{}   & \multicolumn{2}{c}{Predicts}   & \multicolumn{2}{c}{Box$_\textrm{2D}$ gIoU} & \multicolumn{2}{c}{Mask$_\textrm{2D}$ IoU} \\
	Model   & L & S  & Input & Views  & Input & Views\\
	\hhline
	Random depth & \no & \no & 1.00 & 0.36 & 0.73 & 0.39 \\ 
	Fixed depth & \no & \no & 1.00 & 0.34 & 0.73 & 0.38 \\ 
	\methodfive  & \yes & \yes & 1.00 & \bf{0.61} & \bf{0.84} & \bf{0.61}  \\ 
  \end{tabular}
  } 
  \vspace{-2mm}
   \caption{Results on ScanNet val. We compare our model trained with 5 views, \methodfive, to a \emph{random} and \emph{fixed-depth} baseline.}
  \label{tab:scannet_abl}
  \vspace{-5mm}
\end{table}

\section{Discussion}
\label{sec:discussion}

This paper presents a method for predicting 3D object shape and layout from a single image.
We learn without ground truth shapes and layouts, and instead rely on multiple views with 2D annotations.
Our experiments on three datasets show compelling results on images of novel scenes with many objects.
We compare to fully supervised methods and prove our model's superiority to generalize to complex scenes.
On complex scenes at test time, we notice that our model captures layout well but shapes are of lower quality. 
This is not a surprise as supervision comes from 2D silhouettes and thus our 3D shapes cannot easily capture fine details.
Texture information could improve shapes, which we leave for future work.
While our work is a first attempt at learning scene layouts and 3D object shapes from videos end-to-end and without 3D supervision, there is still a lot to be done for models to work in the wild for thousands of object classes and all types of real world scenes.
\section*{Appendix}
\appendix
In this appendix we provide additional information about our approach.
We provide details about the model's architecture and the training recipe and add more qualitative results. 
Video animations which complement our visualizations in the paper can be found in our project's page \url{https://gkioxari.github.io/usl/}.

\section{Comparison to Total3D}
\label{app:total3d}
\begin{figure}[t!]
  \centering
  \scalebox{0.8}{
    { \small
    \begin{tabular}{C{.2\linewidth}C{.4\linewidth}C{.4\linewidth}}
    \hspace{-9mm} Input Image & \hspace{-1.5mm} Total3D~\cite{Nie_2020_CVPR} & \hspace{3mm} \method \\
    \end{tabular}
    } 
    } 
  \includegraphics[width=1.0\linewidth]{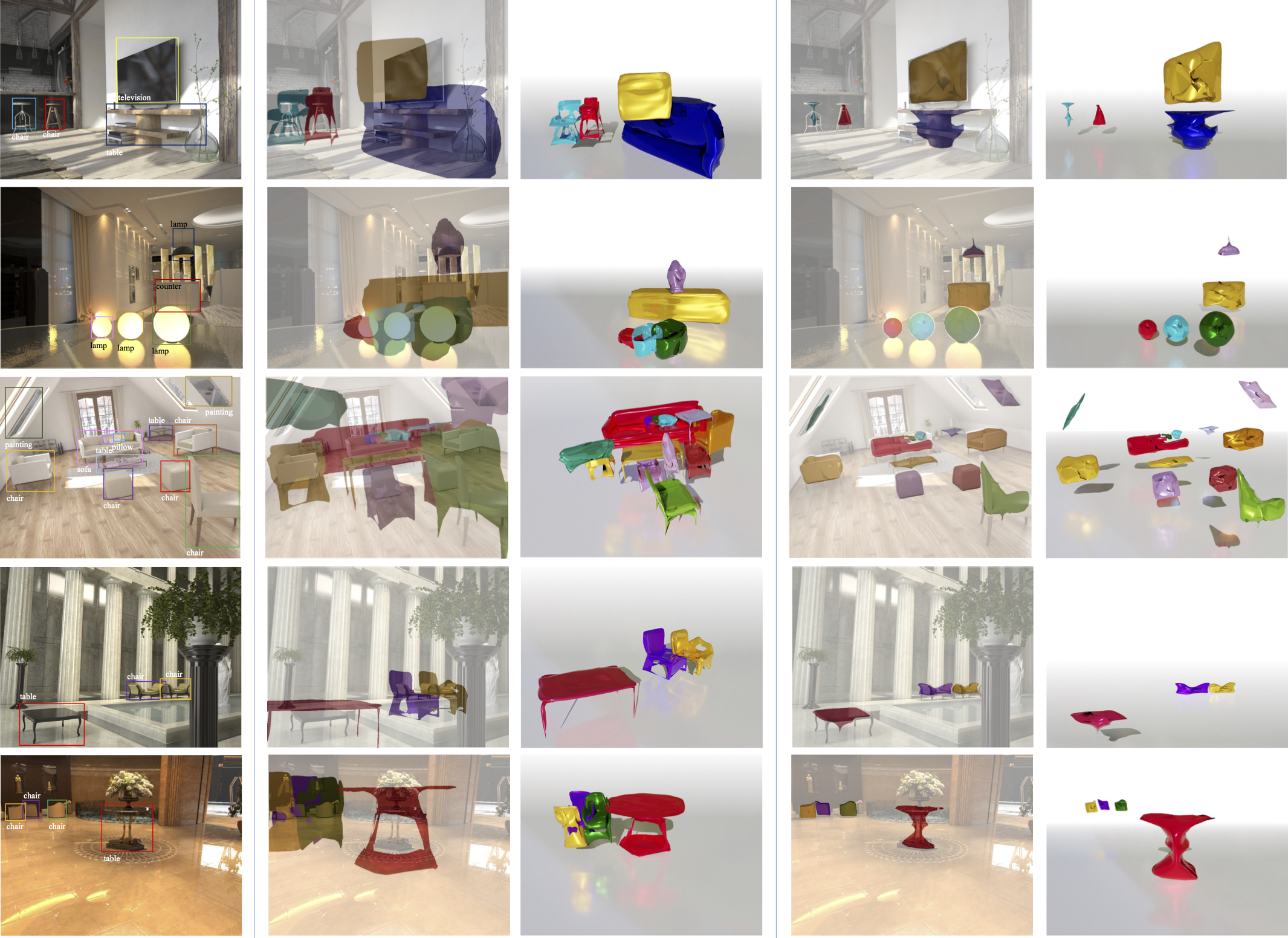}
  \vspace{-6mm}
   \caption{Additional comparisons of Total3D~\cite{Nie_2020_CVPR} and our approach. The input image is shown in the 1$^\textrm{st}$~col. We show the predicted 3D shapes and layout perspectively projected into the image plane along with a 3D visualization of the predicted layout, for Total3D (2$^\textrm{nd}$ \& 3$^\textrm{rd}$~col) and our approach (4$^\textrm{th}$ \& 5$^\textrm{th}$~col).}
   \label{fig:comparison_more}
   \vspace{-4mm}
\end{figure}

Figure~\ref{fig:comparison_more} shows more examples comparing Total3D~\cite{Nie_2020_CVPR} and our approach.
In both the Supplementary and in the main paper, we compare on examples which cover a variety of object and scene types which allows us to better assess generalization of both methods to diverse scenes. 
From Figure~\ref{fig:comparison_more} we notice that when projecting Total3D predictions on the image plane the renderings are not aligned to the input image (2$^\textrm{nd}$ col.). 
This is a consequence of Total3D's model design which regresses to 3D object boxes and shapes without guaranteed alignment with the image. 
In contrast, our predictions are by design aligned to the input image (4$^\textrm{th}$ col.). 
In addition and as noted in the main paper, Total3D has a difficult time placing objects in correct configurations leading to wrong layout and often to large shape intersections (3$^\textrm{rd}$ col.).
In contrast, our approach, even though not trained with 3D supervision, produces more accurate layout predictions (5$^\textrm{th}$ col.).
Finally, we notice that Total3D tends to fetch the nearest shape in semantic space for each object which is commonly not an accurate representation for the particular instance of that object, \eg~the ottoman chairs in the 3$^\textrm{rd}$ example which have a cuboid shape are represented as chairs with four legs and a back (placed under the coffee table).
Our approach more accurately represents object shapes, even if not fine in detail.

\section{Experiments on Scene-Shapes}
\label{app:multishapenet}
We create Scene-Shapes, a dataset of scenes composed of synthetic 3D objects.
We provide details and sample images of the Scene-Shapes dataset.
We show more results comparing to Mesh R-CNN~\cite{gkioxari2019mesh}, trained with 3D supervision on the Scene-Shapes training set.

\subsection{The dataset}
\begin{figure}[t!]
 \centering
    \includegraphics[width=0.4\linewidth]{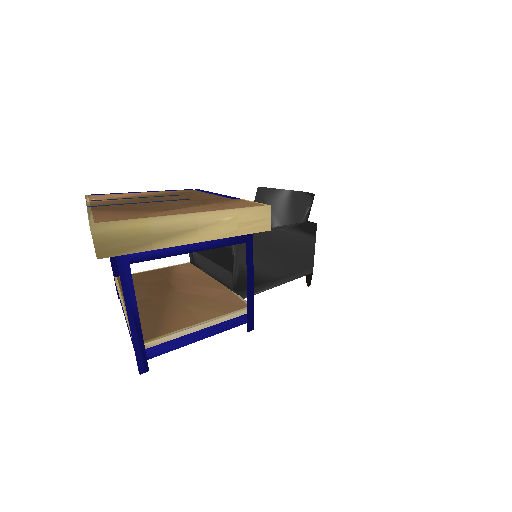}
    \includegraphics[width=0.4\linewidth]{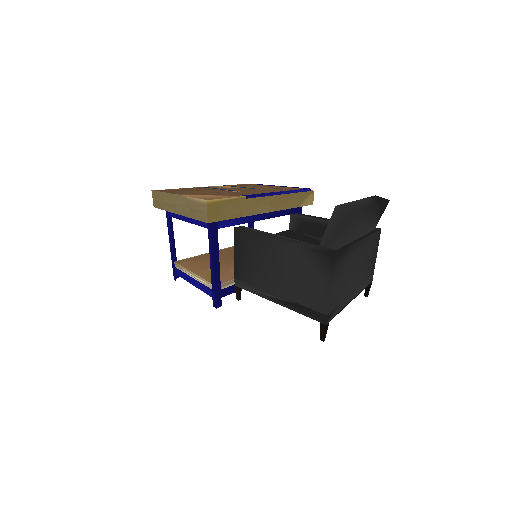}
    \includegraphics[width=0.4\linewidth]{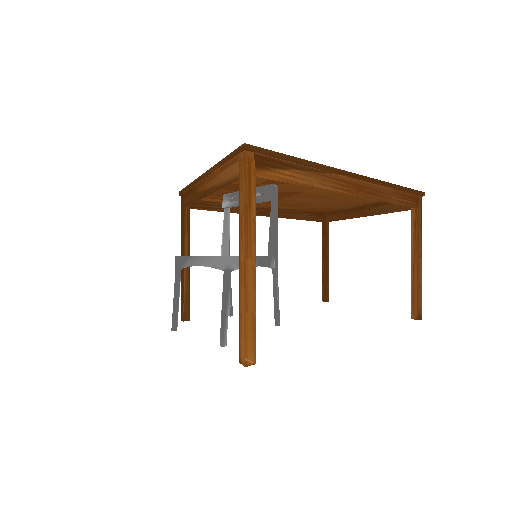}
    \includegraphics[width=0.4\linewidth]{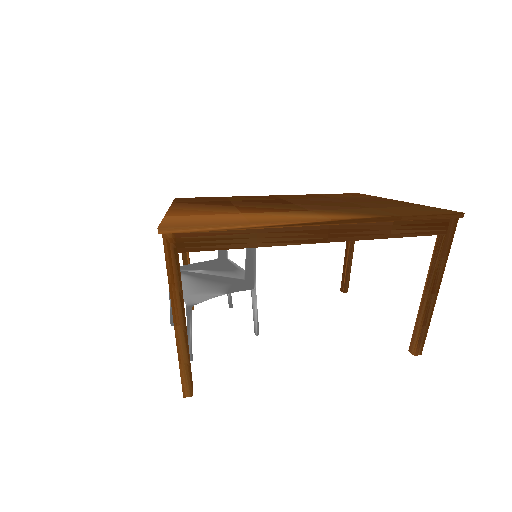}
    \includegraphics[width=0.4\linewidth]{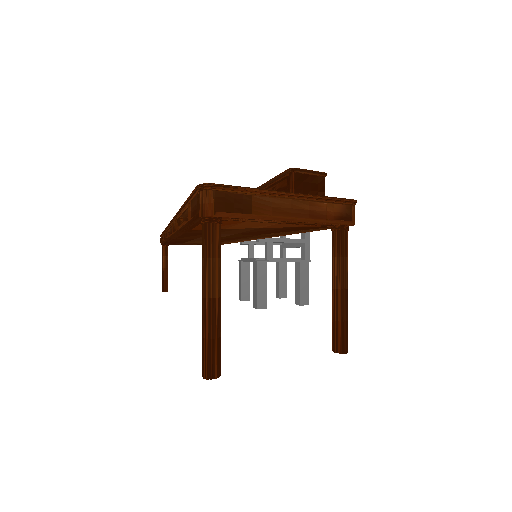}
    \includegraphics[width=0.4\linewidth]{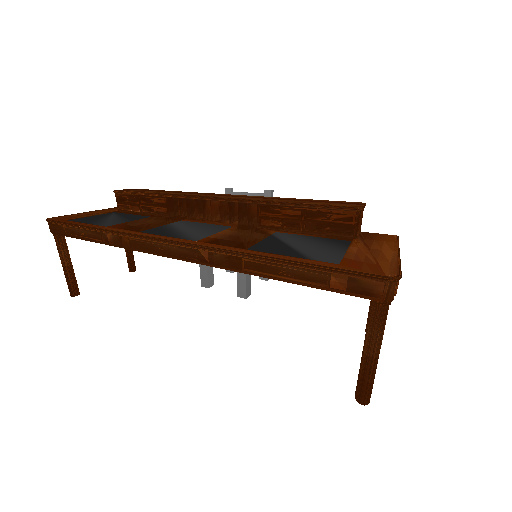}
    \includegraphics[width=0.4\linewidth]{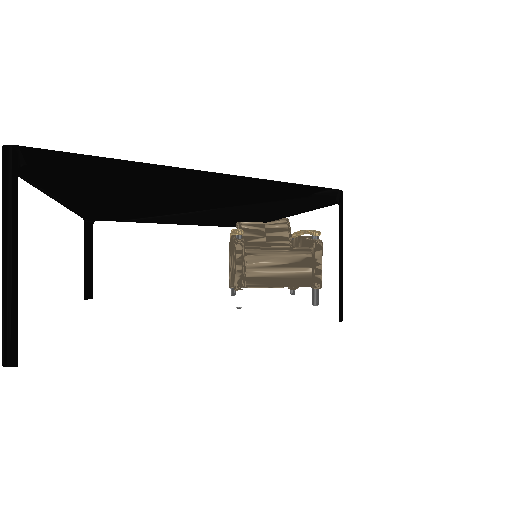}
    \includegraphics[width=0.4\linewidth]{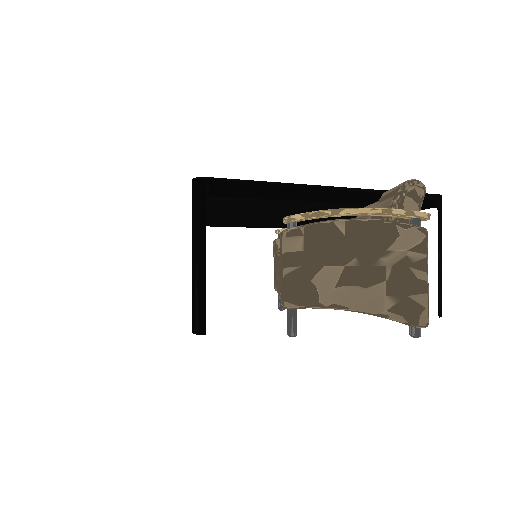}
     \includegraphics[width=0.4\linewidth]{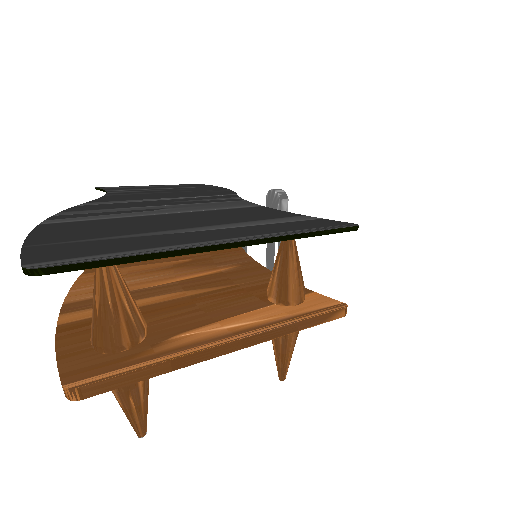}
    \includegraphics[width=0.4\linewidth]{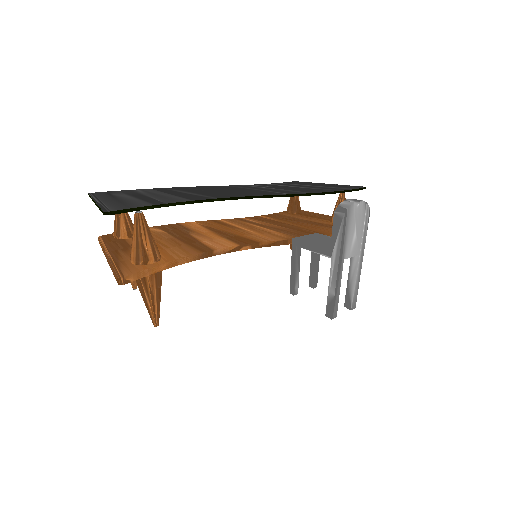}
    \vspace{-2mm}
    \caption{Sample images from the Scene-Shapes dataset. Each row contains an example from two different views. Images are of scenes of object pairs under random spatial configurations.}
    \label{fig:multishapenet_samples}
     \vspace{-2mm}
\end{figure}

As mentioned in the paper, we create the Scene-Shapes dataset from synthetic 3D Warehouse~\cite{warehouse3d} objects by pairing object instances from the \emph{\{chair, table, sofa\}} categories to create scenes.
Specifically, for each scene, we randomly choose a model for an object type. Each model is randomly rotated around the Y axis (``up"), $\theta_Y \in [0^o, 360^o]$, and their center is placed at a random $ (X, Z)$ location on the $Y=0$ plane such that $Z \in [1.5, 1.9]$ and $X \in [-0.4, 0.4]$.
We render the scene by rotating a camera around the objects at multiple azimuth angles and heights. 
We don't exclude scenes with intersecting objects to make the task more challenging as it leads to bigger occlusions, a characteristic of the real world which makes recognition difficult.
Figure~\ref{fig:multishapenet_samples} shows image examples from the dataset. Each row shows two views of the same scene.
The spatial configuration of the pair of objects and their 3D shapes vary across the dataset.

\begin{figure}[t!]
  \centering
  \scalebox{0.99}{
    \begin{tabular}{C{.3\linewidth}C{.31\linewidth}C{.31\linewidth}}
    \hspace{-8mm} Input Image & \hspace{-8mm}\method & \hspace{-12mm}Mesh R-CNN~\cite{gkioxari2019mesh}
    \end{tabular}
    } 
   \includegraphics[width=0.99\linewidth]{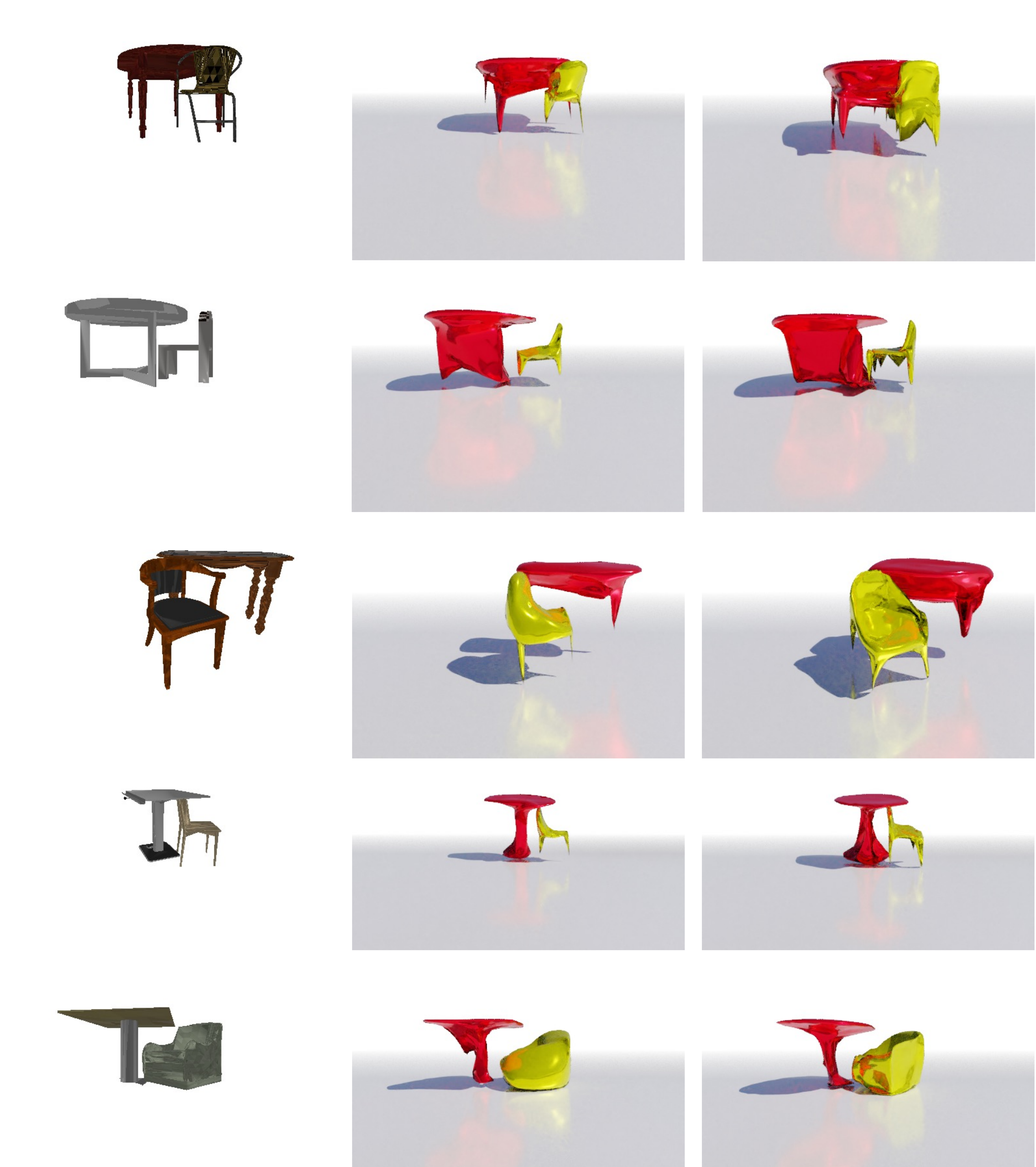}
   \vspace{-3mm}
   \caption{Predictions on Scene-Shapes val. We show the input image (left) and the predicted 3D objects and layout of our \method (middle) and 3D supervised Mesh R-CNN~\cite{gkioxari2019mesh} (right).}
   \label{fig:multishapenet_more}
   \vspace{-3mm}
\end{figure}

\subsection{More results} 

Figure~\ref{fig:multishapenet_more} shows more comparisons of our \method and Mesh R-CNN~\cite{gkioxari2019mesh} on images from Scene-Shapes val. 
Mesh R-CNN is trained with 3D supervision on the Scene-Shapes training set, while \method does not use 3D supervision.
Mesh R-CNN predicts more accurate shapes, as expected.
Even though our \method was not trained with 3D supervision, its predictions are quite good both for estimating the layout and the 3D object shapes. 
This is also validated by our quantitative analysis in Table 1 in the main paper, where our \method performs competitively.
Animations of our predictions are shown in  \url{https://gkioxari.github.io/usl/}.

\subsection{Network architecture and stats}

For the Scene-Shapes experiments we follow exactly the same architecture as Mesh R-CNN~\cite{gkioxari2019mesh} for ShapeNet, without the voxel head as we don't have 3D supervision, and the same training recipe.
Refer to~\cite{gkioxari2019mesh} for more details. 

\mypar{Loss.}
We define the training objective as $\mathcal{L} = \mathcal{L}_\textrm{3D} +\mu_\textrm{reg} \cdot \mathcal{L}_\textrm{reg}$.
We set $\mu_\textrm{reg} = 0.1$ while $\mathcal{L}_\textrm{reg} = \mathcal{L}_\textrm{edge}$ is an edge length regularizer, as in \cite{gkioxari2019mesh}.

\mypar{Training stats.}
We distribute training across 8 V100s with a total of 64 images per batch (8 images per batch per GPU).
For our $5$-view model, each iteration takes 3.4sec and consumes 7.3GB of memory per GPU.
Inference runs at 8fps.

\section{Experiments on Hypersim}
\label{app:hypersim}
\subsection{Video Animations}
Animations of our 3D shape and layout predictions on Hypersim val and test can be found in \url{https://gkioxari.github.io/usl/}.

\subsection{Network architecture and stats}
Table~\ref{tab:hyp_network} shows the network architecture used for our model on Hypersim~\cite{hypersim}. 
We skip the architecture of the RPN and box head, as they are identical to Mask R-CNN~\cite{he2017maskrcnn}. 

\mypar{Loss.}
We define the training objective as $\mathcal{L} = \mathcal{L}_\textrm{2D} + \mu_\textrm{3D} \cdot \mathcal{L}_\textrm{3D} +\mu_\textrm{reg} \cdot \mathcal{L}_\textrm{reg}$.
We set $\mu_\textrm{3D} = 1.0$, $\mu_\textrm{reg} = 0.05$ while $\mathcal{L}_\textrm{reg} = \frac{1}{2} ||dV||_2^2$ is a simple L2 regularizer on the predicted deformations $dV$ (before the \texttt{tanh} layer).
We found that a L2 regularizer performs similarly to $\mathcal{L}_\textrm{edge}$ but the L2 is more efficient allowing us to scale to many object detections, as is the case in Hypersim images.

\mypar{Training stats.}
We distribute training across 8 V100s with a total of 16 images per batch (2 images per batch per GPU).
For our $5$-view model, each iteration takes about 1.1 sec and consumes 6.1GB of memory per GPU. 
We train for 80k iterations.
During inference, our model runs at 3fps. 

\subsection{Performance on the test set.}
For all of our ablations and experiments in the main paper we train on the Hypersim training set and evaluate on the Hypersim validation set. 
Table~\ref{tab:hypersim_test} shows results of our \methodfive on the Hypersim test set with a model trained with 5 views on the Hypersim trainval set.

\section{Experiments on ScanNet}
\label{app:scannet}
\subsection{Network architecture and stats}
Table~\ref{tab:scan_network} shows the network architecture used for our model on ScanNet~\cite{dai2017scannet}. 
We again skip the RPN and box head, as they are identical to Mask R-CNN~\cite{he2017maskrcnn}. 

\mypar{Loss.}
We define the training objective as $\mathcal{L} = \mathcal{L}_\textrm{2D} + \mu_\textrm{3D} \cdot \mathcal{L}_\textrm{3D} +\mu_\textrm{reg} \cdot \mathcal{L}_\textrm{reg}$.
We set  $\mu_\textrm{3D} = 1.0$ and $\mu_\textrm{reg} \cdot \mathcal{L}_\textrm{reg} = 0.05 \cdot \frac{1}{2} ||dV||_2^2 + 0.1 \cdot \mathcal{L}_\textrm{laplac}$, where $ \mathcal{L}_\textrm{laplac}$ is a Laplacian regularizer. 
The laplacian regularizer is added for the ScanNet dataset because the object views are not as diverse compared to Hypersim.
The laplacian regularizer encourages shapes to be smooth and discourages them from degenerating; though is not critical and a similar result could have been achieved perhaps if we increased the weight of the L2 regularizer.

\mypar{Training stats.}
We distribute training across 8 V100s with a total of 16 images per batch (2 images per batch per GPU).
For our $5$-view model, each iteration takes about 0.7 sec and 3.2GB of memory per GPU. 
We train for 200k iterations.

Table~\ref{tab:recipe} summarizes the differences in the training recipes across the three datasets. 
Note that these differences are motivated by (a) fair comparison with prior work and (b) memory/runtime.
On Scene-Shapes our optimizer, backbone, initialization, and regularization are chosen to match the Shapes experiments from Mesh R-CNN~\cite{gkioxari2019mesh}.
Changing these to match Hypersim / ScanNet would not significantly affect results, but hinders fair comparison with~\cite{gkioxari2019mesh}.
The choice of regularizer is not critical; all perform similarly.
We use $\mathcal{L}_{edge}$ on Scene-Shapes to match~\cite{gkioxari2019mesh}, but found that L2 gives similar results with reduced runtime;
adding Laplacian on ScanNet gives slightly smoother meshes but is not critical.
We vary training iterations and LR decay schedule with dataset size.
3 vs 1 refinement stage marginally improves results, but is slower (1.1s vs 0.7s per iteration); due to ScanNet's large size we use 1 stage for faster experiments.

\begin{table}[h!!]
  \centering
  \scalebox{0.85}{
  \begin{tabular}{l|ccc}
    & Scene-Shapes & Hypersim & ScanNet \\
    \hline
    Backbone & R50 & R50-FPN & R50-FPN \\
    Initialization & ImageNet & COCO & COCO \\
    Optimizer / LR & Adam / $10^{-3}$ & SGD / $10^{-2}$ & SGD / $10^{-2}$ \\
    Regularization & $\mathcal{L}_\textrm{edge}$ & $\mathcal{L}_{2}$ & $\mathcal{L}_{2}$ + $\mathcal{L}_\textrm{laplac}$ \\
    Render Resolution & 128 & 72 & 72 \\
    Refinement Stages & 3 & 3 & 1 \\
    Training Iterations & 30K & 80K & 200K
  \end{tabular}}
  \vspace{-3mm}
  \caption{Recipe differences between datasets.}
  \vspace{-5mm}
  \label{tab:recipe}
\end{table}

\begin{table*}
  \centering
  \scalebox{0.99}{
  \begin{tabular}{L{.18\textwidth}|C{.07\textwidth}C{.05\textwidth}|C{.06\textwidth}C{.06\textwidth}C{.06\textwidth}C{.06\textwidth}C{.06\textwidth}C{.06\textwidth}}
  	\toprule
	\multicolumn{1}{c}{}   & \multicolumn{2}{c}{Predicts}   & \multicolumn{2}{c}{Box$_\textrm{2D}$ gIoU} & \multicolumn{2}{c}{Mask$_\textrm{2D}$ IoU} & \multicolumn{2}{c}{Depth $L_1$ ($\downarrow$)}  \\
	Model   & Layout & Shape  & Input & Views  & Input & Views & Input & Views \\
	\hhline
	\methodfive & \yes & \yes & 1.00 & 0.35 & 0.69 & 0.36 & 2.07 & 1.98 \\ 
	\methodfive w/ detections & \yes & \yes & 0.93  & 0.34 & 0.68 & 0.35 & 2.08 & 2.00  \\ 
  \end{tabular}
  } 
  \vspace{-3mm}
   \caption{Performance of our \methodfive on the Hypersim test set trained with 5 views on Hypersim trainval, with ground truth input boxes (1$^\textrm{st}$ row) and the model's object detections (2$^\textrm{nd}$ row).}
   \label{tab:hypersim_test}
   \vspace{-3mm}
\end{table*}

\begin{table*}
  \centering
  \setlength{\tabcolsep}{1mm}
  \scalebox{0.99}{
  \begin{tabular}{|c|c|l|c|}
    \hline
    \textbf{Index} & \textbf{Inputs} & \textbf{Operation} & \textbf{Output shape} \\
    \hline
    (1) & Input & Input Image & $H \times W \times 3$ \\
    (2) & (1) & Backbone: ResNet-50-FPN p2 level & $\frac{H}{4} \times \frac{W}{4} \times 256$ \\
    (3) & (2) & RoIAlign: Pools and Averages RoI features & $1 \times 256$ \\
    (4) & (3) & $4 \times$ Linear(256, 256) &  $1\times 256$ \\
    (5) & (4) & Layout $z$: Linear(256, $|\mathcal{C}|$) & $1\times |\mathcal{C}|$ \\
    (6) & (4) & Layout $\rho$: Linear(256, $|\mathcal{C}|$) & $1\times |\mathcal{C}|$ \\
    (7) & (2) & Stage 1: RoIMap: Pools features from ico3 sphere & $|V|\times256$, $|F|\times3$ \\ 
    (8) & (7) & Stage 1: $3\times$ GraphConv(256 + 3, 256) &  $|V|\times256$, $|F|\times3$ \\
    (9) & (8) & Stage 1: Linear(256 + 3, 3) &  $|V|\times3$, $|F|\times3$ \\
    (10) & (2), (9) & Stage 2: RoIMap: Pools features from (9) & $|V|\times256$, $|F|\times3$ \\ 
    (11) & (10) & Stage 2: $3\times$ GraphConv(256 + 3, 256) &  $|V|\times256$, $|F|\times3$ \\
    (12) & (11) & Stage 2: Linear(256 + 3, 3) &  $|V|\times3$, $|F|\times3$ \\
    (13) & (2), (12) & Stage 3: RoIMap: Pools features from (12) & $|V|\times256$, $|F|\times3$ \\ 
    (14) & (13) & Stage 3: $3\times$ GraphConv(256 + 3, 256) &  $|V|\times256$, $|F|\times3$ \\
    (15) & (14) & Stage 3: Linear(256 + 3, 3) &  $|V|\times3$, $|F|\times3$ \\
    \hline
  \end{tabular}}
  \caption{
    Overall architecture of our model on Hypersim. The backbone, RPN and box branches are identical to Mask R-CNN~\cite{he2017maskrcnn}.
    The RPN produces a bounding box prediction for anchors at each spatial location in the input feature map;
    a subset of these candidate boxes are processed by the other branches, but here we show only the shapes resulting from
    processing a single box for the subsequent task-specific heads. Here, $|\mathcal{C}|$ is the number of categories.
  }
  \vspace{-3mm}
  \label{tab:hyp_network}
\end{table*}

\begin{table*}[h!]
  \centering
  \setlength{\tabcolsep}{1mm}
  \scalebox{0.99}{
  \begin{tabular}{|c|c|l|c|}
    \hline
    \textbf{Index} & \textbf{Inputs} & \textbf{Operation} & \textbf{Output shape} \\
    \hline
    (1) & Input & Input Image & $H \times W \times 3$ \\
    (2) & (1) & Backbone: ResNet-50-FPN p2 level & $\frac{H}{4} \times \frac{W}{4} \times 256$ \\
    (3) & (2) & RoIAlign: Pools and averages RoI features & $1 \times 256$ \\
    (4) & (3) & $4 \times$ Linear(256, 256) &  $1\times 256$ \\
    (5) & (4) & Layout $C_Z$: Linear(256, $|\mathcal{C}|$) & $1\times |\mathcal{C}|$ \\
    (6) & (4) & Layout $\rho_Z$: Linear(256, $|\mathcal{C}|$) & $1\times |\mathcal{C}|$ \\
    (7) & (2) & Stage 1: RoIMap: Pools features from ico3 sphere & $|V|\times256$, $|F|\times3$ \\ 
    (8) & (7) & Stage 1: $3\times$ GraphConv(256 + 3, 256) &  $|V|\times256$, $|F|\times3$ \\
    (9) & (8) & Stage 1: Linear(256 + 3, 3) &  $|V|\times3$, $|F|\times3$ \\
 \hline
  \end{tabular}}
  \caption{
    Overall architecture of our model on ScanNet. The backbone, RPN and box branches are identical to Mask R-CNN~\cite{he2017maskrcnn}.
    The RPN produces a bounding box prediction for anchors at each spatial location in the input feature map;
    a subset of these candidate boxes are processed by the other branches, but here we show only the shapes resulting from
    processing a single box for the subsequent task-specific heads. Here, $|\mathcal{C}|$ is the number of categories.
  }
  \vspace{-3mm}
  \label{tab:scan_network}
\end{table*}

{\small
\bibliographystyle{ieee_fullname}
\bibliography{references}

\begin{thebibliography}{10}\itemsep=-1pt

\bibitem{Bao2013}
Sid~Yingze Bao, Manmohan Chandraker, Yuanqing Lin, and Silvio Savarese.
\newblock Dense object reconstruction with semantic priors.
\newblock In {\em CVPR}, 2013.

\bibitem{blanz1999}
Volker Blanz and Thomas Vetter.
\newblock A morphable model for the synthesis of 3d faces.
\newblock In {\em SIGGRAPH}, 1999.

\bibitem{chen2016single}
Weifeng Chen, Zhao Fu, Dawei Yang, and Jia Deng.
\newblock Single-image depth perception in the wild.
\newblock In {\em NeurIPS}, 2016.

\bibitem{chen2019learning}
Wenzheng Chen, Huan Ling, Jun Gao, Edward Smith, Jaakko Lehtinen, Alec
  Jacobson, and Sanja Fidler.
\newblock Learning to predict 3d objects with an interpolation-based
  differentiable renderer.
\newblock In {\em NeurIPS}, 2019.

\bibitem{choy2016r2n2}
Christopher~B. Choy, Danfei Xu, JunYoung Gwak, Kevin Chen, and Silvio Savarese.
\newblock 3{D}-{R}2{N}2: A unified approach for single and multi-view 3d object
  reconstruction.
\newblock In {\em ECCV}, 2016.

\bibitem{dai2017scannet}
Angela Dai, Angel~X. Chang, Manolis Savva, Maciej Halber, Thomas Funkhouser,
  and Matthias Nie{\ss}ner.
\newblock Scannet: Richly-annotated 3d reconstructions of indoor scenes.
\newblock In {\em CVPR}, 2017.

\bibitem{dai2017bundlefusion}
Angela Dai, Matthias Nie{\ss}ner, Michael Zollh{\"o}fer, Shahram Izadi, and
  Christian Theobalt.
\newblock Bundlefusion: Real-time globally consistent 3d reconstruction using
  on-the-fly surface reintegration.
\newblock {\em ToG}, 2017.

\bibitem{dame2013}
Amaury Dame, Victor~A. Prisacariu, Carl~Y. Ren, and Ian Reid.
\newblock Dense reconstruction using 3d object shape priors.
\newblock In {\em CVPR}, 2013.

\bibitem{eigen2014depth}
David Eigen, Christian Puhrsch, and Rob Fergus.
\newblock Depth map prediction from a single image using a multi-scale deep
  network.
\newblock In {\em NeurIPS}, 2014.

\bibitem{fan2017point}
Haoqiang Fan, Hao Su, and Leonidas~J Guibas.
\newblock A point set generation network for 3d object reconstruction from a
  single image.
\newblock In {\em CVPR}, 2017.

\bibitem{felzenszwalb2009object}
Pedro~F Felzenszwalb, Ross~B Girshick, David McAllester, and Deva Ramanan.
\newblock Object detection with discriminatively trained part-based models.
\newblock {\em PAMI}, 2009.

\bibitem{felzenszwalb2012distance}
Pedro~F Felzenszwalb and Daniel~P Huttenlocher.
\newblock Distance transforms of sampled functions.
\newblock {\em Theory of computing}, 2012.

\bibitem{gkioxari2019mesh}
Georgia Gkioxari, Jitendra Malik, and Justin Johnson.
\newblock Mesh {R}-{CNN}.
\newblock In {\em ICCV}, 2019.

\bibitem{ucmrGoel20}
Shubham Goel, Angjoo Kanazawa, , and Jitendra Malik.
\newblock Shape and viewpoints without keypoints.
\newblock In {\em ECCV}, 2020.

\bibitem{gupta2013perceptual}
Saurabh Gupta, Pablo Arbelaez, and Jitendra Malik.
\newblock Perceptual organization and recognition of indoor scenes from rgb-d
  images.
\newblock In {\em CVPR}, 2013.

\bibitem{han2020drwr}
Zhizhong Han, Chao Chen, Yu-Shen Liu, and Matthias Zwicker.
\newblock Drwr: A differentiable renderer without rendering for unsupervised 3d
  structure learning from silhouette images.
\newblock {\em arXiv preprint arXiv:2007.06127}, 2020.

\bibitem{hartley2003multiple}
Richard Hartley and Andrew Zisserman.
\newblock {\em Multiple view geometry in computer vision}.
\newblock Cambridge university press, 2003.

\bibitem{he2017maskrcnn}
Kaiming He, Georgia Gkioxari, Piotr Doll\'{a}r, and Ross Girshick.
\newblock {Mask R-CNN}.
\newblock In {\em ICCV}, 2017.

\bibitem{he2016deep}
Kaiming He, Xiangyu Zhang, Shaoqing Ren, and Jian Sun.
\newblock Deep residual learning for image recognition.
\newblock In {\em CVPR}, 2016.

\bibitem{hane2014}
Christian Häne, Nikolay Savinov, and Marc Pollefeys.
\newblock Class specific 3d object shape priors using surface normals.
\newblock In {\em CVPR}, 2014.

\bibitem{warehouse3d}
Trimble Inc.
\newblock 3d warehouse.
\newblock https://3dwarehouse.sketchup.com/.

\bibitem{kanazawa2018learning}
Angjoo Kanazawa, Shubham Tulsiani, Alexei~A Efros, and Jitendra Malik.
\newblock Learning category-specific mesh reconstruction from image
  collections.
\newblock In {\em ECCV}, 2018.

\bibitem{kar2017learning}
Abhishek Kar, Christian H{\"a}ne, and Jitendra Malik.
\newblock Learning a multi-view stereo machine.
\newblock In {\em NeurIPS}, 2017.

\bibitem{kato2018neural}
Hiroharu Kato, Yoshitaka Ushiku, and Tatsuya Harada.
\newblock Neural 3{D} mesh renderer.
\newblock In {\em CVPR}, 2018.

\bibitem{kendall2017end}
Alex Kendall, Hayk Martirosyan, Saumitro Dasgupta, Peter Henry, Ryan Kennedy,
  Abraham Bachrach, and Adam Bry.
\newblock End-to-end learning of geometry and context for deep stereo
  regression.
\newblock In {\em ICCV}, 2017.

\bibitem{kirillov2019pointrend}
Alexander Kirillov, Yuxin Wu, Kaiming He, and Ross Girshick.
\newblock {PointRend}: Image segmentation as rendering.
\newblock In {\em CVPR}, 2020.

\bibitem{kulkarni2020acsm}
Nilesh Kulkarni, Abhinav Gupta, David Fouhey, and Shubham Tulsiani.
\newblock Articulation-aware canonical surface mapping.
\newblock In {\em CVPR}, 2020.

\bibitem{kulkarni2019csm}
Nilesh Kulkarni, Abhinav Gupta, and Shubham Tulsiani.
\newblock Canonical surface mapping via geometric cycle consistency.
\newblock In {\em ICCV}, 2019.

\bibitem{li2018differentiable}
Tzu-Mao Li, Miika Aittala, Fr{\'e}do Durand, and Jaakko Lehtinen.
\newblock Differentiable monte carlo ray tracing through edge sampling.
\newblock {\em TOG}, 2018.

\bibitem{li2020self}
Xueting Li, Sifei Liu, Kihwan Kim, Shalini~De Mello, Varun Jampani, Ming-Hsuan
  Yang, and Jan Kautz.
\newblock Self-supervised single-view 3d reconstruction via semantic
  consistency.
\newblock In {\em ECCV}, 2020.

\bibitem{li2018megadepth}
Zhengqi Li and Noah Snavely.
\newblock Megadepth: Learning single-view depth prediction from internet
  photos.
\newblock In {\em CVPR}, 2018.

\bibitem{lin2017feature}
Tsung-Yi Lin, Piotr Doll{\'a}r, Ross Girshick, Kaiming He, Bharath Hariharan,
  and Serge Belongie.
\newblock Feature pyramid networks for object detection.
\newblock In {\em CVPR}, 2017.

\bibitem{lin2014microsoft}
Tsung-Yi Lin, Michael Maire, Serge Belongie, James Hays, Pietro Perona, Deva
  Ramanan, Piotr Doll{\'a}r, and C~Lawrence Zitnick.
\newblock {Microsoft COCO}: Common objects in context.
\newblock In {\em ECCV}, 2014.

\bibitem{liu2019soft}
Shichen Liu, Weikai Chen, Tianye Li, and Hao Li.
\newblock Soft rasterizer: Differentiable rendering for unsupervised
  single-view mesh reconstruction.
\newblock In {\em ICCV}, 2019.

\bibitem{loper2014opendr}
Matthew~M Loper and Michael~J Black.
\newblock Open{DR}: An approximate differentiable renderer.
\newblock In {\em ECCV}, 2014.

\bibitem{luo2020consistent}
Xuan Luo, Jia-Bin Huang, Richard Szeliski, Kevin Matzen, and Johannes Kopf.
\newblock Consistent video depth estimation.
\newblock {\em TOG}, 2020.

\bibitem{Silberman:ECCV12}
Pushmeet~Kohli Nathan~Silberman, Derek~Hoiem and Rob Fergus.
\newblock Indoor segmentation and support inference from rgbd images.
\newblock In {\em ECCV}, 2012.

\bibitem{Nie_2020_CVPR}
Yinyu Nie, Xiaoguang Han, Shihui Guo, Yujian Zheng, Jian Chang, and Jian~Jun
  Zhang.
\newblock Total3dunderstanding: Joint layout, object pose and mesh
  reconstruction for indoor scenes from a single image.
\newblock In {\em CVPR}, 2020.

\bibitem{nimier2019mitsuba}
Merlin Nimier-David, Delio Vicini, Tizian Zeltner, and Wenzel Jakob.
\newblock Mitsuba 2: a retargetable forward and inverse renderer.
\newblock {\em TOG}, 2019.

\bibitem{novotny2019c3dpo}
David Novotny, Nikhila Ravi, Benjamin Graham, Natalia Neverova, and Andrea
  Vedaldi.
\newblock C3dpo: Canonical 3d pose networks for non-rigid structure from
  motion.
\newblock In {\em ICCV}, 2019.

\bibitem{ravi2020pytorch3d}
Nikhila Ravi, Jeremy Reizenstein, David Novotny, Taylor Gordon, Wan-Yen Lo,
  Justin Johnson, and Georgia Gkioxari.
\newblock Accelerating 3d deep learning with pytorch3d.
\newblock {\em arXiv:2007.08501}, 2020.

\bibitem{ren2015faster}
Shaoqing Ren, Kaiming He, Ross Girshick, and Jian Sun.
\newblock {Faster R-CNN}: Towards real-time object detection with region
  proposal networks.
\newblock In {\em NeurIPS}, 2015.

\bibitem{Rezatofighi_2018_CVPR}
Hamid Rezatofighi, Nathan Tsoi, JunYoung Gwak, Amir Sadeghian, Ian Reid, and
  Silvio Savarese.
\newblock Generalized intersection over union.
\newblock In {\em CVPR}, 2019.

\bibitem{hypersim}
Mike Roberts, Jason Ramapuram, Anurag Ranjan, Atulit Kumar, Miguel~Angel
  Bautista, Nathan Paczan, Russ Webb, and Joshua~M. Susskind.
\newblock {Hypersim}: {A} photorealistic synthetic dataset for holistic indoor
  scene understanding.
\newblock In {\em ICCV}, 2021.

\bibitem{scharstein2002taxonomy}
Daniel Scharstein and Richard Szeliski.
\newblock A taxonomy and evaluation of dense two-frame stereo correspondence
  algorithms.
\newblock {\em IJCV}, 2002.

\bibitem{schmidt2017self}
Tanner Schmidt, Richard Newcombe, and Dieter Fox.
\newblock Self-supervised visual descriptor learning for dense correspondence.
\newblock In {\em IEEE Robotics and Automation Letters}, 2017.

\bibitem{smith2019geometrics}
Edward~J Smith, Scott Fujimoto, Adriana Romero, and David Meger.
\newblock Geometrics: Exploiting geometric structure for graph-encoded objects.
\newblock In {\em ICML}, 2019.

\bibitem{song2015sun}
Shuran Song, Samuel~P Lichtenberg, and Jianxiong Xiao.
\newblock Sun rgb-d: A rgb-d scene understanding benchmark suite.
\newblock In {\em CVPR}, 2015.

\bibitem{pix3d}
Xingyuan Sun, Jiajun Wu, Xiuming Zhang, Zhoutong Zhang, Chengkai Zhang, Tianfan
  Xue, Joshua~B Tenenbaum, and William~T Freeman.
\newblock Pix3d: Dataset and methods for single-image 3d shape modeling.
\newblock In {\em CVPR}, 2018.

\bibitem{factored3dTulsiani17}
Shubham Tulsiani, Saurabh Gupta, David Fouhey, and Alexei A. Efrosand~Jitendra
  Malik.
\newblock Factoring shape, pose, and layout from the 2d image of a 3d scene.
\newblock In {\em CVPR}, 2018.

\bibitem{wang2018pixel2mesh}
Nanyang Wang, Yinda Zhang, Zhuwen Li, Yanwei Fu, Wei Liu, and Yu-Gang Jiang.
\newblock Pixel2{M}esh: Generating 3{D} mesh models from single {RGB} images.
\newblock In {\em ECCV}, 2018.

\bibitem{wu2019detectron2}
Yuxin Wu, Alexander Kirillov, Francisco Massa, Wan-Yen Lo, and Ross Girshick.
\newblock Detectron2.
\newblock \url{https://github.com/facebookresearch/detectron2}, 2019.

\bibitem{ye2021shelf}
Yufei Ye, Shubham Tulsiani, and Abhinav Gupta.
\newblock Shelf-supervised mesh prediction in the wild.
\newblock In {\em CVPR}, 2021.

\bibitem{yin2021learning}
Wei Yin, Jianming Zhang, Oliver Wang, Simon Niklaus, Long Mai, Simon Chen, and
  Chunhua Shen.
\newblock Learning to recover 3d scene shape from a single image.
\newblock In {\em CVPR}, 2021.

\bibitem{zhang2020perceiving}
Jason~Y Zhang, Sam Pepose, Hanbyul Joo, Deva Ramanan, Jitendra Malik, and
  Angjoo Kanazawa.
\newblock Perceiving 3d human-object spatial arrangements from a single image
  in the wild.
\newblock In {\em ECCV}, 2020.

\bibitem{zhou2017unsupervised}
Tinghui Zhou, Matthew Brown, Noah Snavely, and David~G Lowe.
\newblock Unsupervised learning of depth and ego-motion from video.
\newblock In {\em CVPR}, 2017.

\end{thebibliography}
}

\end{document}